# Inductive Inference in Supervised Classification

Ali Amiryousefi
Department of Mathematics & Statistics
University of Helsinki,
FI-00014, Finland
ali.amiryousefi@helsinki.fi
Supervisor: Prof. Jukka Corander

**Abstract**

Inductive inference in the supervised classification context constitutes to methods and approaches to assign some objects or items into different predefined classes using a formal rule that is derived from training data and possibly some additional auxiliary information. The optimality of such an assignment varies under different conditions due to intrinsic attributes of the objects being considered for such a task. One of these cases is when all the objects' features are discrete variables with *a priori* known categories. As another example, one can consider a modification of this case with *a priori* unknown categories. These two cases are the main focus of this thesis and based on Bayesian inductive theories, de Finetti type exchangeability is a suitable assumption that facilitates the derivation of classifiers in the former scenario. On the contrary, this type of exchangeability is not applicable in the latter case, instead, it is possible to utilize the partition exchangeability due to John Kingman. These two types of exchangeabilities are discussed and furthermore here we investigate inductive supervised classifiers based on both types of exchangeabilities. We further demonstrate that the classifiers based on de Finetti type exchangeability can optimally handle test items independently of each other in the presence of infinite amounts of training data while on the other hand, classifiers based on partition exchangeability still continue to benefit from joint labeling of all the test items. Additionally, it is shown that the inductive learning process for the simultaneous classifier saturates when the amount of test data tends to infinity.

# Contents

# Chapter 1

# Introduction

Over the past century, numerous arguments have been put forward to support the view that in a world where due to our incomplete knowledge uncertainty is dominant, inductive inference should be Bayesian that provides us with necessary tools to make judgement with available information. However, the Bayesian perspective is far from universally accepted and many other approaches to inductive reasoning have had a peaceful coexistence with it over several decades [1],[43],[51]. One of the intriguing cases of inductive inference is when we want to construct rules for classifying some given items into different given classes. With the assumption of 'given classes', we are implicitly confining our definition to supervised classification in which we emphasise the fact that we are aware of all possible categories that an item can be assigned to. Hence we do not need to consider any new classes in addition to our prior classes to encompass one or more items in the future - denoting the semi supervised classification definition. Even in this very case, we can narrow down our attention to the cases where those features that we want to make the classification rules based on, are discrete. In this case one can consider two distinct scenarios: (I) the possible discrete values are known *a priori* or (II) not known *a priori*, meaning that it is possible that in the future we come across some *new* values for our features that we have not considered before. These cases are the main focus of this thesis and in chapter 3 and 5 we will discuss methods to tackle these specific cases (I) and (II), respectively. While in (I) case the de Finetti type arguments represent fruitful and convenient assumptions, they do not apply to data of the (II) case, and consequently it is less intuitive how a coherent Bayesian inductive classifier could look like when unpredictable events can occur. One feasible answer to this difficulty lies in the theory of exchangeable random partitions developed by Sir John Kingman[1] [38-41].

[1]See also the elegant discussion of its appearances within several fields of study and its inductive implications by Sandy Zabell [53].

In a statistical classification task we make a distinction between two types of data that our classification machinery will be based on, namely training and test data. Training data is a set of items for which, labels are known and we train our machine with them. On the other hand, test data is a set of data for which we are not aware of the labels, but instead, we need to use observed features of the items to predict them in a probabilistic fashion. The goal of classification is to find some suitable way to assign each item of the test data set to the optimal class among the existing alternatives. What constitutes an optimal solution, varies considerably due to many factors such as: potential necessity of classifying the items immediately upon their arrival sequentially, possibility of waiting to pile the whole test data and classify them simultaneously, allowance of reclassifying the items upon arrival of the new items, etc. Here we restrict our attention only on simultaneous and sequential learning, precluding reclassification, and construct our theory based on the de Finetti type of exchangeability and the partition exchangeability. With the use of the latter, we will arrive at an inductive supervised classifier as a generalization of more standard classifiers arising under the ordinary exchangeability. In particular, we demonstrate that classifiers based on partition exchangeability behave differently from classifiers derived under de Finetti type exchangeability. Using the result from [13],[14], we also demonstrate that marginal and simultaneous predictive classifiers of test data become congruent almost surely when the amount of training data tends to infinity. In contrary, under partition exchangeability, a positive probability always remains for the event that marginal and simultaneous predictive classifiers do not coincide, such that the population of test data items will play a clear role when assessing how surprising any particular observed event is, as noted in [11].

In the second chapter we present the preliminary theory needed to tackle the classification task with the Bayesian approach. The concepts of prior, likelihood, posterior and prediction are addressed in this chapter. The third chapter discusses firstly the principles of Bayesian predictive classification in (I) case with emphasis on the cases that we are provided with abundantly either test or training data. Secondly, the asymptotic behaviour of the classifiers of the two cases are stated under two main theorems. Chapter four discusses the concept of exchangeability both in its de Finetti form and its alternative due to Kingman. Chapter five serves as the complementary counterpart of the third chapter in the (II) scenario wherein a theorem exhibiting the asymptotic behaviour of the predictive classifiers under partition exchangeability when the amount of training data tends to infinity is stated. Challenges and arguments are discussed in the sixth chapter, and lastly, appendices present necessary background results on mathematical functions and statistical distributions, as well as some technical details related to the presented results.

# Chapter 2

# Bayesian Predictive Inference

The ultimate goal of the Bayesian approach can be stated as to completely model a set of observables and then calculate the probability distribution of the unobserved quantities of interest conditional on those already observed as discussed by Geisser in [25], in which he extensively provides treatments for predictive inference. With only a few exceptions, achieving this goal seems in general to be too burdensome a task. Hence, for the most part we use the conventional Bayesian paradigm that models distributions of observables given parameters (or index set). One then can assign either subjective prior densities whenever possible (or appropriate) or useful so-called noninformative prior densities for the parameters of the statistical model. Once this is done, inferences of decisions about unobserved values conditional on those observed can be obtained from calculable probability distributions.

In this chapter we recapitulate the theory of the Bayesian prediction as the cornerstone of the following chapters. One can refer to [7] for extensive theory of the contents. The first section will discuss the basic elements of the Bayesian theory, namely prior, likelihood, and posterior as tools that enable us to start the machinery of the Bayesian predictive inference which is thoroughly investigated in the consecutive section.

## 2.1 Elements of Bayesian Theory

Bayesian approach in statistics was developed after Bayes' work (1763) with the use of what is known as the Bayes' formula which is nothing but a noble way of applying the basic rules of probability, namely sum and product. Virtually in all the problems tackled with Bayesian methodology, one needs to formulate three fundamental components namely likelihood function, prior, and posterior. While prior synthesises our uncertainty before observing any outcome, likelihood function is encompassing the information obtained from what we have observed throughout our course of experiment, and finally posterior (The Holy Grail) characterizes our degree of belief after merging the two former components. First we will consider the likelihood and its derivation from probability function. Then, the prior and its different forms will be placed under investigation while posterior is presented lastly.

### 2.1.1 Likelihood

To begin with, we note that likelihood function in general is the same as the probability function of observables regarded as a function of parameters rather than a function of random variables. With this consideration, now suppose that $X$ is a random variable with either finite or infinite, countable or uncountable state space (support) for which probability distribution is given by $f(X = x|\theta)$. We further suppose that this probability distribution is normalized, meaning that it sum to one when summed over all possible values of $x \in S_x$ where $S_x$ is the support of the $X$. Although this condition is not strictly needed from the computational perspective in the Bayesian paradigm ([48]), we will find it comfortable to develop our theory with such an assumption while we reserve the notation $f^*(.|.)$ for unnormalized version of our distribution that is otherwise the same. One can think of functional form of $f(.|.)$ as measure of belief or probability regarding the defined random variable. In general, knowing this functional form will enable us to answer questions related to our random variable in any form. No matter what is our enquiry, as long as it is about our random variable $X$, we can have a deterministic probability(!) as an answer. A one simple example is the case of Bernoulli[1] random variable that is defined with $S_x = \{0, 1\}$. Consider 1 as success and 0 as failure. One can define

$$f(X = x|\theta) = \theta^x(1-\theta)^{1-x}, \quad x \in S_x. \tag{2.1}$$

Using this function, we can compute the probability of questions related to our random variable. For example, probability of $X = 0$ according to above formula is $1-\theta$. We can even generalize our question beyond single query about different values of random variable and consider investigation of the probability of a sequence of random variables following a specific distribution function $f(.|.)$. Answering to these questions is still tractable with referring to the functional format of our distribution, but the exception is that we need to know more information about the process which is giving rise to a specific sequence of the random variables. This new information incorporates a structure in our outcome probabilities. There are infinitely many cases that one can conjecture about the way that in each step we obtain our next value of the random variable for a given size of a sequence. The simplest case would be s process of independent random variables, meaning that the next value of our random variable is not affected by the previous nor none of the other values. Another highly applicable case is the Markov chain process that relates the values of the future to some order of earlier values, such that the length of this relevant subprocess is the order of the Markov chain. Note that the above case with complete independence is a specific case of this process, *i.e.* a zero order Markov chain. To make it more precise, we consider the previous scenario with Bernoulli experiment. Now suppose that we want to answer the question about the probability of the sequence $\{X_1 = 1, X_2 = 1, X_3 = 0\}$. Subscripts are reflecting the fact that we are interested in the sequence of the random variables in time. We use $N$ to denote the length of the sequence, hence $N = 3$. If we agree that each of the variables is following the same Bernoulli distribution defined in (2.1), based on their independence we can write this probability

(2.2)
$$f(X_1 = 1, X_2 = 1, X_3 = 0|\theta) = f(X_1 = 1|\theta)f(X_2 = 1|\theta)f(X_3 = 0|\theta) = \theta^2(1-\theta).$$

[1]For the mathematical format of some common distributions see the Appendix C.

Here, we are differentiating between all the possible cases that one can hypothesise about the combinations of the values that our random variable can take. In this case with easy calculation we find that there is $2^3 = 8$ cases [2] that are considered to be different from each other. The triplet $(X_1, X_2, X_3)$ denote the values of the same random variable in time, such that ordering of the subscript holds. Then, all different eight cases of this vector are as

$$(X_1, X_2, X_3) \mapsto \Sigma_{i=1}^{N} X_i = n \tag{2.3}$$

$$\begin{aligned} (0,0,0) &\mapsto 0 \\ (1,0,0),(0,1,0),(0,0,1) &\mapsto 1 \\ (1,1,0),(1,0,1),(0,1,1) &\mapsto 2 \\ (1,1,1) &\mapsto 3. \end{aligned}$$

Probability of all these cases can be calculated the same way as in (2.2) as long as the independence condition holds and the ordering of the values are important. With modifying the conditions mentioned above, we arrive at different probabilities for each case. For instance by suppressing the time factor, we get an indifference of cases aligned horizontally in (2.3) (each line). In this way we shift our attention from the exact triplet $(X_1, X_2, X_3)$ to the sum of its components that is denoted as $n$ for each different case in (2.3).Now our question can be postulate with regard to the number of successes $n$, in a known sequence of random variables which is a sufficient statistic. This shift of attention from the triplet itself to its more compressed version information *i.e.* sufficient statistic, will impose an unavoidable change in our probability structure regarding the outcomes. More specifically, with these considerations we will evaluate the probability of having two successes in our three Bernoulli experiments as

$$
\begin{aligned}
(2.4)\quad f(\Sigma_{i=1}^{3}X_i = 2|\theta) = f(n = 2|\theta) &= \sum_{x_i \in S_x : n=2} f(X_1 = x_1, X_2 = x_2, X_3 = x_3|\theta) \\
&= f(X_1 = 1, X_2 = 1, X_3 = 0|\theta) \\
&+ f(X_1 = 1, X_2 = 0, X_3 = 1|\theta) \\
&+ f(X_1 = 0, X_2 = 1, X_3 = 1|\theta),
\end{aligned}
$$

by the use of (2.2) formula we obtain the above equality as

$$
f(\Sigma_{i=1}^{3}X_i = 2|\theta) = f(n = 2|\theta) = 3(\theta^2(1-\theta)). \tag{2.5}
$$

[2] In general, this value is $N$ times of the size of the support of $X$, *i.e.* $|S_x|^N$.

Notice that we considered the case with only two possible outcomes so we needed at least frequency of one of the states in a case of *known* size of a sequence. So more precisely we can consider $\mathbf{n} = (n_1, n_2)$ as the frequency vector such that $n_1$ and $n_2$ are the frequencies of the different states of the Bernoulli process that sum to size of the sequence $N$. So in our case we had $\mathbf{n} = (1, 2)$ comprising to $(X_1 = 1, X_2 = 0, X_3 = 1)$, $(X_1 = 1, X_2 = 1, X_3 = 0)$, and $(X_1 = 0, X_2 = 1, X_3 = 1)$.

As noted earlier, when we construct our probability function for a random variable, one can think of this function as a function of a parameter.

$$
f(X|\theta), \quad \theta \in \Theta.
$$

This is a function that answers the question about the degree of plausibility of a data in a probabilistic manner for a given parameter. In other words, how *likely* it is to observe (a sequence of) a random variable when we know the parameter (hence *likelihood*). So we tend to refer to these quantities such as what is represented in (2.2) and (2.5) as likelihood functions that can take different values based on different parameter $\theta$ input. We further refer to all the possible values that $\theta$ can take as parameter space and denote it by $\Theta$.

### 2.1.2 Prior

In the previous section we derived the likelihood function as a function of the parameter, which is affecting all of our answers to the probabilistic questions. We also denoted all the possible values of this parameter as $\Theta$. One can introduce a distribution for the parameter over its support and denote this by

$$\pi(\theta), \quad \theta \in \Theta.$$

This prior[3] distribution can express our uncertainty about the parameter *before* observing the outcomes of the experiment (hence *prior*). In another words, we have to formulate our (prior) belief about the parameter using a mathematical function such as $\pi(\theta)$. This will enable us to answer the queries about the parameter before even getting any knowledge about the outcome of an experiment related to $\theta$ through its functional form.

There is an abundant literature around how one should incorporate one's prior knowledge about the parameter in the probability realm and a comprehensive treatment of the issue is given in [34]. Also one may further want to differentiate between the cases of objective and subjective Bayesian approach as in [4],[27], respectively. In the case of insufficient reason or indifference about the parameter one may opt for the simplest *non-informative* prior. It is commonly applied when there is no knowledge indicating unequal probabilities. This type of prior is what is used and emphasized in the objective Bayesian approach. In contrast, an alternative approach would be to choose an *informative* prior that would be based on careful examination of expert knowledge about the parameter $\theta$ and elicitation of a prior distribution from an expert or a group of experts. This is what is referred to as subjective Bayesian approach in the statistics area. If the data are very informative about the quantity being estimated, then an uninformative prior is a quick and easy choice. Otherwise, it is suggested to choose an informative prior that is incorporating this available piece of information to our inference. The effect of different priors is considered in the following section but here, we would not carry the burden of such a differentiation about our prior as long as its mathematical format is known explicitly. Hence we only focus on our prior as a tool to help us proceed with our inference and we ignore how it has been formulated.

### 2.1.3 Posterior

So far we have built the foundations of Bayesian inference by deriving likelihood and prior functions. Now the posterior is the what is helping us with our queries about the parameter *after* observing the outcomes of the experiment(hence *posterior*). Both likelihood and prior will affect our posterior. In fact, if the data are extremely informative, then nearly any prior would lead to the same posterior distribution. But if the data are sparse, then the posterior will be heavily influenced by the prior and it is more important to think how the prior was chosen and how sensitive the result is to different priors. This can be better understood by looking at the mathematical format of the posterior. With the previous notation, define our prior as

$$\pi(\theta), \quad \theta \in \Theta, \tag{2.6}$$

and our likelihood as

$$f(X|\theta), \quad \theta \in \Theta. \tag{2.7}$$

[3]The correct way of showing this is $\pi(\theta|\delta)$ that $\delta$ is denoting the hyperparameters of the prior distribution, but we tend to stop here and suppress this notation at this stage.

Then one can make use of the sum and product rule of probability and derive the famous Bayes formula as

$$\pi(\theta|X) = \frac{f(X|\theta)\pi(\theta)}{f(X)}, \quad \theta \in \Theta. \tag{2.8}$$

That the denominator of the fraction is the marginal distribution of the data that can be derived by summing (or integrating) the joint distribution $f(X, \theta)$ over the parameter space. Irrelevance of this marginal distribution to the inference about the parameter will allow us to often suppress it in our notation which will lead us to

$$\pi(\theta|X) \propto f(X|\theta)\pi(\theta), \quad \theta \in \Theta. \tag{2.9}$$

This is particularly useful when we use the so-called conjugate prior [7]. That is, when we express our prior knowledge using such a function then combined with the likelihood, posterior will remains always in the same class of mathematical functions as the prior with respect to teh parameter. Recall $f^*(.|.)$ that was the unnormalized version of $f(.|.)$. In this setting we also can make use of this alternative to our convenience. Besides, in this case, the use of unnormalized or even improper priors [4] is allowed. We denote these cases with $\pi^*(.)$. To make it precise, we will revisit our example from section 2.1.1 where the likelihood function is given by (2.2). We further assume an uninformative case or indifference for our parameter as prior, and choose our prior as $Unif(0, 1)$ *i.e.* the standard

[4]Improper prior is a prior that is not sum (or integrate) to any finite value, hence not normalizable. Nevertheless, in some cases it will lead to proper posterior that justifies its use for certain purposes.

Uniform distribution[5]. For simplicity[6], we further denote our data as $X$ such that $\sum_{i=1}^{3} X_i = 2 \equiv X$.

$$\pi^*(\theta) = c, \quad \theta \in (0,1) \quad \& \quad c \in \mathbb{R}.$$

And according to (2.2), our likelihood is

$$f^*(X|\theta) = \theta^2(1-\theta), \quad \theta \in (0,1),$$

that will lead to our posterior as

$$\pi(\theta|X) \propto \theta^2(1-\theta), \quad \theta \in (0,1).$$

One can recognize this as the kernel of the Beta distribution. Hence our posterior is given by

$$\pi(\theta|X) = \frac{\Gamma(3+2)}{\Gamma(3)\Gamma(2)}\theta^{3-1}(1-\theta)^{2-1}, \quad \theta \in (0,1).$$

This is the case of conjugacy between the Beta and Binomial distribution that gives us a closed form of the Beta distribution as our posterior whenever we express our prior distribution and likelihood with Beta and Binomial distributions, respectively. After obtaining the posterior, one can answer all the possible probabilistic questions regarding the parameter, such as the probability that a parameter falls in a given interval, etc. A comprehensive consideration of both Bayesian methods and diverse use of them in real world is given in [26] while [48] thoroughly investigate methods to numerically handle those posterior functions that are hard to evaluate explicitly.

## 2.2 Bayesian Prediction

In previous section we have considered how to obtain the posterior based on prior and likelihood. This is indeed useful, since it equips us with a tool that can help us with our queries about the parameter in our model. While posterior density summarized our current uncertainty about an unknown quantity, predictions of future experiments and events could sometimes be even more interesting. As discussed in [25] this is often the ultimate purpose of modelling. With this emphasize on prediction we tend to shift our attention to the values of our random variable in the consecutive run of the experiment and we are seeking a systematic way to formalize our uncertainty about possible values that a random variable could attain. We further want to have a concrete probabilistic statement about those values, hence one can talk about the predictive distribution function.

[5]Standard Uniform distribution is the special case of the Beta distribution with its parameters equal with one, *i.e.* $\theta \sim Beta(1,1) \equiv \theta \sim Unif(0,1)$.

[6]This abuse of notation hold only through out this section.

In previous section, we were concerned solely about the parameter. We discussed synthesizing our knowledge into a mathematical form as prior, further we conditioned our likelihood on a given (known) parameter, and finally we derived our posterior as our updated function encompassing uncertainty about parameter. But in prediction area on the other hand, one needs not to think of the parameter as such. Since here we are only concerned about the future value of the random variable given a history of that random variable so far. Of course for derivation of such a probabilistic function about random variable we need to inject and relate our functions to the parameter, however, its not necessary though, to have a physical interpretation of the parameter. To make it clear, lets consider the case of Bernoulli experiment introduced earlier. Suppose that we are not aware about the real value of the $\theta$ parameter (that is often the case in real phenomena) and we have executed the experiment already $N$ times which from these, $X$ times of them turned out to be success[7]. Now we can derive the prior predictive probability for the first experiment to have a success turn-out as

$$p(X_1 = 1) = \int_0^1 p(X_1 = 1, \theta) d\theta,$$

which is equal to

$$p(X_1 = 1) = \int_0^1 p(X_1 = 1|\theta)\pi(\theta) d\theta = \int_0^1 \theta\pi(\theta) d\theta = E(\theta), \tag{2.10}$$

that $E(.)$ is the expectation function and $\pi(.)$ is the prior distribution of $\theta$. This formula tells us that the prior predictive probability of observing the value of the first experiment as success without any other information (observing other variables) is simply the expectation of the prior distribution of the parameter itself.

For the case of posterior predictive probability that is the probability of observing the next random variable to be success while we have the history variables $x_1, x_2 \ldots x_N$ that among them $X$ numbers were successes, we have

$$p(X_{N+1} = 1|X, N) = \int_0^1 p(X_{N+1} = 1, \theta|X, N) d\theta,$$

that again with the use of the product rule of probability we obtain

$$p(X_{N+1} = 1|X, N) = \int_0^1 p(X_{N+1} = 1|\theta)\pi(\theta|X, N) d\theta = E(\theta|X, N), \tag{2.11}$$

[7] Note that in harmony with our frequency vector introduced in the Likelihood section this information is the same as $\mathbf{n} = (X, N - X)$.

where $\pi(.|.)$ is the posterior of the $\theta$ and we have used the fact $X_{N+1} = 1$ is independent from $N$ and $X$ given $\theta$. This formula is expressing our post uncertainty about the value of the random variable after observing a history of that variable that is shown to be equal as expectation of the posterior probability distribution of the parameter.

In special case of assigning the $Beta(\alpha, \beta)$ distribution as the prior, (2.10) and (2.11) will turn out to be $\frac{\alpha}{\alpha+\beta}$ and $\frac{X+\alpha}{N+\alpha+\beta}$, respectively. This beautifully shows us the way that our uncertainty is updated as we go along the course of the experiment. One need not to make any assumption on the parameter or know it in advance. As long as the goal is prediction, we can learn by looking at our observations and make our probabilistic statements more precise. It is also notable that, updating probabilities at once or sequentially will not make any difference in our probabilities formulation. One can think of $N$ as 1 in each step and update the probabilities, then after observing the related value of the random variable, one updates the posterior and plugs it in (2.11) to obtain the probability of the next step and so on.

Next, one can consider the predictive distribution. To make it precise consider an experiment of repeating the Bernoulli trials (as considered before) for $N$ times and denote the number of the successes by $X$. If we know the parameter $\theta$, then $X \sim Bin(N, \theta)$. But in case we are not aware about the value of $\theta$ we have to integrate over all conceivable values of that over its space $\Theta$ to obtain the *predictive* distribution of the $X$ such as

$$p(X|N) = \int_0^1 p(X, \theta|N) d\theta = \int_0^1 p(X|N, \theta)\pi(\theta) d\theta, \tag{2.12}$$

that in the specific case of assigning the prior as $Beta(\alpha, \beta)$ and further $A = X + \alpha, B = N - X + \beta$ we will have

(2.13)

$$\begin{aligned} p(X|N) &= \int_0^1 \frac{\Gamma(N+1)}{\Gamma(X+1)\Gamma(N-X+1)} \frac{\Gamma(\alpha+\beta)}{\Gamma(\alpha)\Gamma(\beta)} \theta^{X+\alpha-1}(1-\theta)^{N-X+\beta-1} d\theta \\ &= \frac{\Gamma(N+1)\Gamma(\alpha+\beta)}{\Gamma(X+1)\Gamma(N-X+1)\Gamma(\alpha)\Gamma(\beta)} \int_0^1 \theta^{X+\alpha-1}(1-\theta)^{N-X+\beta-1} d\theta, \end{aligned}$$

recognizing the last part of the second equality as the kernel of the $Beta(A, B)$

distribution and also using the so-called beta function[8] we can write

$$
\begin{aligned}
p(X|N) &= \frac{\Gamma(N+1)\Gamma(\alpha+\beta)}{\Gamma(X+1)\Gamma(N-X+1)\Gamma(\alpha)\Gamma(\beta)} \frac{\Gamma(A)\Gamma(B)}{\Gamma(A+B)} \\
&= \binom{N}{X} \frac{\Gamma(\alpha+\beta)\Gamma(A)\Gamma(B)}{\Gamma(A+B)\Gamma(\alpha)\Gamma(\beta)} \\
&= \binom{N}{X} \frac{beta(X+\alpha, N-X+\beta)}{beta(\alpha,\beta)}.
\end{aligned} \tag{2.14}
$$

This predictive distribution of $X$ is said to be *Beta-Binomial* distribution. It is widely emphasized for instance in the food safety microbial risk assessments to describe the number of contaminated servings $X$ among $N$ servings, under uncertainty about the true fraction $\theta$ of contaminated servings in a large (infinite) population ([9],[10]). In risk assessment literature, the conditional distribution of $X$ (Binomial) is often called as the variability distribution of $X$, and the distribution of $\theta$ (Beta) as the uncertainty distribution. Hence, it is often said in that 'variability and uncertainty are separated'. In Bayesian context, both distributions are expressions of uncertainty and Beta-Binomial distribution reflects both uncertainties. This can be either prior or posterior predictive distribution.

As another example of deriving the predictive distribution, consider the previously described scenario but with only difference that each experiment is governed by different parameter $\theta$. So in each step the outcome of experiment is Binomial $Bin(1, \theta_i)$. Our uncertainty about all $\theta_i$ is assumed to be described as some prior distribution $\pi(\theta_i)$ for $i = 1, 2, ..., N$(which can be $Beta(\alpha, \beta)$ as we assumed in following derivation). Now seeking the $X$ distribution as the total number of successes in our course of experiment we find

$$
p(X|N) = \int_0^1 \cdots \int_0^1 p(X|\theta_1, \ldots, \theta_N)\pi(\theta_1, \ldots, \theta_N) d\theta_1 \cdots d\theta_N
$$

$$
= \int_0^1 \cdots \int_0^1 \binom{N}{X} \prod_{i=1}^{X} \theta_{k_i} \prod_{i=N-X}^{N} (1-\theta_{k_i}) \prod_{i=1}^{N} \frac{\Gamma(\alpha+\beta)}{\Gamma(\alpha)\Gamma(\beta)} \theta_{k_i}^{\alpha-1}(1-\theta_{k_i})^{\beta-1} d\theta_{k_1} \cdots d\theta_{k_N},
$$

here, $k_1, \ldots, k_N$ is some permutation of the indices $i$. After rearranging the terms

[8]For mathematical account of this function see the Appendix B.

in this expression, we get

$$P(X|N) = \binom{N}{X} \int_0^1 \cdots \int_0^1 \prod_{i=1}^{X} \frac{\Gamma(\alpha+\beta)}{\Gamma(\alpha)\Gamma(\beta)} \theta_{k_i}^{\alpha+1-1} (1-\theta_{k_i})^{\beta-1}$$

$$\times \prod_{i=N-X}^{N} \frac{\Gamma(\alpha+\beta)}{\Gamma(\alpha)\Gamma(\beta)} \theta_{k_i}^{\alpha-1} (1-\theta_{k_i})^{\beta+1-1} d\theta_{k_1} \cdots d\theta_{k_N},$$

and then by integrating over each $\theta_i$ one by one, we get

$$p(X|N) = \binom{N}{X} E(\theta_i)^X E(1-\theta_i)^{N-X} = Bin(N, \frac{\alpha}{\alpha+\beta}). \tag{2.15}$$

This is a distribution that depends on $N$ and the expected value of $\theta_i$, so the prior distribution of $\theta_i$ affects the result via its expected value only.

Note that these two predictive posteriors that we have derived in (2.14) and (2.15) will exhibit their difference for example in the case that we are sampling from a closed population for which we are ignorant about the prevalence of a certain attribute of its elements, and the case (for assessing some questions about the same attribute) where we are sampling from different populations (or the dynamic one with regard to prevalence of that attribute) where neither the prevalence parameter values, nor their possible equality is known.

Of course the set of predictive distribution by no means is limited to the above examples but still one other interesting form of this distribution harnessed exclusively in the classification and machine learning area arise under the Multinomial assumption for the mechanism to consider and quantify the data, accompanied by prior belief about this distribution's parameters expressed in the form of the Dirichlet distribution. In the same way one can think of the *Dirichlet-Multinomial* distribution as the predictive distribution in this case. This particular choice of prior and formulating the likelihood is of particular interest in the bioinformatics literature as considered in [42].

# Chapter 3

# Predictive Classification I

So far we have equipped ourself with the necessary tools which enable us to proceed further with the inductive inference in classification. Of course the use of prediction is far wider to be restricted only to classification, nevertheless, harnessing this concept for the data with discrete (and finite) classes is of particular value in machine learning. In this chapter, we consider generic problem of classification *i.e.*, assigning some items into a discrete set of classes using observed characteristics or features of the items and assessing the uncertainty related to the assignments conditional on all relevant information available. We consider the *supervised* classification[1], where all the potential classes are *a priori* given[2]. While this setting prevails throughout this and fifth chapter, in this chapter we further focus on the case of predictive classification for data with *finite alphabet* or more precisely with *fixed alphabet*, meaning that we only consider the case where all the possible values of features of data are discrete, finite, and known in advance. The special case regarding data with *a priori unknown* alphabets is discussed in chapter five. We further consider the inductive inference behaviour in two different cases distinctly when we are able tp gather more training data and the case that we have finite number of training data but we are able to accumulate more test items. Of course the difference between these two cases is that in former, we are fed with increasingly many known labels as we are gathering more training data, while in latter, we need to learn the labels of the abundant test data that we obtain. Apart from this fact, test and training data can be considered in the same manner in both cases. In supervised classification it is typical to classify test items or samples individually, independently of the classification decisions made for any other item on the test set. This is typically motivated by the generating *i.i.d.* assumption, where the features for any test item are conditionally independent of the features for any other item given the fixed generating probability measure. However, inductive learning theory based on predictive modelling implies that the test items *should* be classified *simultaneously*, because the item data can be considered conditionally independent only given their *joint labelling* with respect to *a priori* specified classes. Marginal dependence exists between the test items in the predictive probability distribution, since the underlying generating probability measure is not exactly known, or in another words, we are ignorant about the true parameter(s). In the second section we will show that this dependence vanishes in the case that the amount of training data tends to infinity, and consequently, the two learning approaches become equivalent. And this is what causes the process of learning to saturate when the amount of test data increases. We further note that the classification principles themselves are universally applicable to any data types, but here we restrict our attention to multivariate discrete data in the derivation of mathematically explicit predictive probabilities. We also consider a predictive classification principle which uses marginal posterior distribution of item labels derived from the simultaneous classifier. Such a classifier is referred to as a *marginalized,* in contrast to the standard *marginal* classifier which treats all test items independently of each other. The marginalized classifier arises under the predictive learning framework for any particular item when the labels of remaining items are considered as nuisance parameters.

[1]For the advances of supervised classification over nearly half a century, see [2], [8],[19], [46], [22], [28], and [5]

[2]This definition implies (but not necessarily confine) the *semi-supervised* classification definition as the case where a certain set of eligible classes is *a priori* determined, while the items are note forced to solely be allocated to such classes, but can also form previously unknown groups during the classification task. For extensive investigation of the both cases in predictive classification see [13].

We already note that this chapter is closely based on [13], who operationalized the idea of simultaneous classification. The structure of this chapter is as follows. In the first section we derive the predictive supervised classification framework, followed by the formulations of predictive probability distributions for the considered classifiers. The asymptotic relationships between simultaneous and marginal classifier when the amount of the training and test data tends to infinity are investigated in the next sections, respectively.

## 3.1 Predictive Classification and Classification Structures

Let $M$ be a set of $m$ training items $i \in M$ for which we observe finite vectors $\mathbf{z}_i$ of $d$ features, such that each element $z_{ij}$ in $\mathbf{z}_i$ belongs to a finite alphabet, $z_{ij} \in \mathcal{X}_j = \{1, \ldots, r_j\}, r_j \geq 1, j = 1, \ldots, d$. The joint feature space is represented

by the Cartesian product

$$\mathbf{z}_i \in \mathcal{X} = \times_{j=1}^{d} \mathcal{X}_j. \tag{3.1}$$

Further let $T$ denote the joint classification, *i.e., labels* of the training data items in $M$, such that each element $T_i \in \{1, \ldots, k\}$ assigns an item $i \in M$ unambiguously to a class indexed by the integer $c$, $1 \leq c \leq k$. The $k$ classes and the labels in $T$ are assumed to have been established *a priori* using some suitable information. The total collection of training feature data is denoted by $\mathbf{z}^{(M)}$ and the subset of training feature data for the class $c$ by $\mathbf{z}^{(M_c)}, M_c \subset M$, with the size $m_c = |M_c|$. Note that $\mathbf{z}^{(M)}$ does not include the labels specified by the training data, as the labels are instead defined by $T$. Let now $N$ be a set of $n$ test items $i \in N$ for which we observe finite feature vectors $\mathbf{x}_i \in \mathcal{X}$ defined in the same space as $\mathbf{z}_i$, such that the total collection of test feature data is denoted as $\mathbf{x}^{(N)}$. For any subset $A \subset N$ we let $\mathbf{x}^{(A)}$ be the corresponding subset of the test feature data. In particular, $N_c \subset N$ will be used for the set of test items assigned to a particular class and $n_c = |N_c|$ is the size of this set.

By a classification structure $S$ we refer to a unique assignment of all the test items in $N$ to a finite set of classes, which is the $k$ classes present in $T$. We let the $\mathcal{S}$ denotes the family of all such structures. The cardinality of $\mathcal{S}$ equals the Cartesian product $\{1, \ldots, k\}^n$ and further each element of $S$ is defined analogously to $T$ such that $S_i \in \{1, \ldots, k\}$.

### 3.1.1 Predictive classifier

We note that the classification principles considered in this section hold more universally for arbitrary data types, but explicit expressions for the predictive probabilities will be given in the next section under Predictive distribution under exchangeability. We consider first the joint conditional distribution of $S$, the training data $\mathbf{z}^{(M)}$ and the test data $\mathbf{x}^{(N)}$, given the fixed labeling $T$. Let $p(S|T) > 0, S \in \mathcal{S}$, be a prior probability for a classification structure $S$ conditional on $T$, which let here be implicitly defined. To define the posterior probability for $S$, we define first the probability of all feature data $(\mathbf{x}^{(N)}, \mathbf{z}^{(M)})$ conditional on $T$ as

$$p(\mathbf{x}^{(N)}, \mathbf{z}^{(M)}|T) = \sum_{S \in \mathcal{S}} p(\mathbf{x}^{(N)}|\mathbf{z}^{(M)}, S, T) p(\mathbf{z}^{(M)}|T) p(S|T), \tag{3.2}$$

where we have used the fact that the features of the training data is independent of the labels of the test data given labels of the training data. *i.e.* $p(\mathbf{z}^{(M)}|T) = p(\mathbf{z}^{(M)}|S, T)$. Here $p(\mathbf{x}^{(N)}|\mathbf{z}^{(M)}, S, T)$ is the (posterior) predictive probability of test data given the training data as noted in [7]. Formally in the case of a parametric[3]

[3]Note that here we refer to all the parameters of the features as $\theta$ . So one should be cautious about accordance of integrals and other mathematical derivations while dealing with it, and

family for the features by $\theta \in \Theta$, the predictive probability will equal an infinite mixture of data distributions over a set of random parameters $\theta$, such that

$$p(\mathbf{x}^{(N)}|\mathbf{z}^{(M)}, S, T) = \int_{\Theta} p(\mathbf{x}^{(N)}|\theta, S, T) dF(\theta|\mathbf{z}^{(M)}, T), \tag{3.3}$$

where $dF(\theta|\mathbf{z}^{(M)}, T)$ is a conditional probability distribution for the model parameters given the training data. This is predictive in the sense that it assigns a probability on future observations by learning about the parameters from the training data as in [25]. It is natural to assume independence between the distribution of the training features data and $S$, whereby the posterior probability of $S$ equals

$$p(S|\mathbf{x}^{(N)}, \mathbf{z}^{(M)}, T) = \frac{p(\mathbf{x}^{(N)}|\mathbf{z}^{(M)}, S, T)p(S|T)}{\sum_{S \in \mathcal{S}} p(\mathbf{x}^{(N)}|\mathbf{z}^{(M)}, S, T)p(S|T)}, \tag{3.4}$$

since the term $p(\mathbf{z}^{(M)}|T)$ cancels out. Also, in the above expression it is implicitly assumed that the training data labelling $T$ is given, such that one does not need to consider a prior distribution $P(T)$.

In an alternative case, one can contemplate a parametric generative model for the features and labels, such that the joint probability of all observables and latent variables at time $t = n$ can be specified as

$$p(\mathbf{x}^{(N)}|S, \theta)p(\mathbf{z}^{(M)}|T, \theta)p(\theta)p(S|\psi)p(T|\psi)p(\psi), \tag{3.5}$$

where the terms are defined as follow: $p(\mathbf{x}^{(N)}|S, \theta)$ is the likelihood of the observed features of test items given the hidden sequence of labels $S$, $p(\mathbf{z}^{(M)}|T, \theta)$ is the likelihood of the observed features of training items given the known sequence of labels $T$, $\theta \in \Theta$ is a finite dimensional parameter, $p(\theta)$ is the prior density of $\theta$ describing uncertainty about the parameters, $p(S|\psi)$ is the probability of the hidden sequence of labels given a generating model indexed by the finite dimensional parameter $\psi \in \Psi$, $p(T|\psi)$ is the corresponding probability of the known sequence of labels, and finally, $p(\psi)$ is a prior density characterizing uncertainty about $\psi$.

A common approach to classification removes the unknown parameters $\theta, \psi$ by replacing them with point estimates $\hat{\theta}, \hat{\psi}$, such as those provided by the maximum likelihood method, and then proceeds as if the generating model was entirely known. However, it is a theoretical fact that such an approach is potentially neglecting a substantial amount of classification uncertainty. A Bayesian predictive should not mix it with the notation introduced in last chapter that was assuming $\theta$ only for the one dimensional case.

classification framework is on the other hand, based on the predictive distribution of the test data, conditional on the known quantities alone, which for $t = n$ equals (3.6)

$$p(\mathbf{x}^{(N)}|\mathbf{z}^{(M)},T) = \frac{\sum_{S\in\mathcal{S}} \int_{\theta\in\Theta} p(\mathbf{x}^{(N)}|S,\theta)p(\mathbf{z}^{(M)}|T,\theta)p(\theta)d\theta \int_{\psi\in\Psi} p(S|\psi)p(T|\psi)p(\psi)d\psi}{\int_{\theta\in\Theta} p(\mathbf{z}^{(M)}|T,\theta)p(\theta)d\theta \int_{\psi\in\Psi} p(T|\psi)p(\psi)d\psi},$$

where each element in the sum $\sum_{S\in\mathcal{S}}$ equals the predictive probability conditional on the hidden sequence of labels (3.7)

$$p(\mathbf{x}^{(N)},\mathbf{z}^{(M)},S,T) = \int_{\theta\in\Theta} p(\mathbf{x}^{(N)}|S,\theta)p(\mathbf{z}^{(M)}|T,\theta)p(\theta)d\theta \int_{\psi\in\Psi} p(S|\psi)p(T|\psi)p(\psi)d\psi.$$

Using Bayes' rule, the above predictive distribution can be converted to the joint posterior over the labels

$$p(S|\mathbf{x}^{(N)},\mathbf{z}^{(M)},T) = \frac{p(\mathbf{x}^{(N)},\mathbf{z}^{(M)},S,T)}{\sum\limits_{S\in\mathcal{S}} p(\mathbf{x}^{(N)},\mathbf{z}^{(M)},S,T)}, \tag{3.8}$$

which is the same as (3.4).

Now with this background, there are many different strategies one can consider for assigning each test item that is available in this stage, to a putative class. But in particular we investigate three different alternatives.

(i) If the $n$ test items are considered independently of each other, we obtain the standard **Marginal Predictive Classifier**. We denote this classifier by MPC. This classifier yields the following posterior distribution for the label of item $i \in N$

$$p(S_i = c|\mathbf{z}^{(M)},\mathbf{x}_i,T) = \frac{p(\mathbf{x}_i|\mathbf{z}^{(M_c)},S_i=c,T)p(S_i=c)}{\sum_{c=1}^{k} p(\mathbf{x}_i|\mathbf{z}^{(M_c)},S_i=c,T)p(S_i=c)}, \quad c = 1,\ldots,k \tag{3.9}$$

where $p(\mathbf{x}_i|\mathbf{z}^{(M_c)})$ is the predictive probability of $\mathbf{x}_i$ in class $c$ and $p(S_i = c)$ is the prior probability of label $c$ and $p(\mathbf{z}^{(M_c)}|T, S_i = c)p(T|S_i = c)$ is cancelled from the numerator and denominator. A predictive classifier that assigns the label $S_i$ by maximizing the posterior probability can be shown to minimize the mean risk of misclassification[4] [46].

(ii) A **Simultaneous Predictive Classifier**, denoted as SPC, assigns jointly the posterior probability on the collection of labels for all items according to (3.4).

[4]As noted in [46], in a non-sequential modelling context, MPC is not an optimal approach to classifying any single test item when data from other test items are present. A proof of optimality of the predictive classification has also been given in [44] by considering the expected loss for the case where training and test items arrive sequentially. For more recent work on the use of Bayesian predictive classification for instance in speech recognition, see [31].

This classifier combines the information from all the previous test items with training data when assessing the label of a new test item[5]. A SPC which assign the labels $S$ by maximizing the posterior (3.4), is optimal under the zero-one loss function

$$L_1(S,\omega) = \begin{cases} 0, & \text{if } \omega = S \\ 1, & \text{if } \omega \neq S \end{cases}, \tag{3.10}$$

which imposes a constant loss on all label sequences $S$ apart from the sequence of true labels $\omega$.

Ripley [46] notes that the marginal classifier does not have as high a prediction accuracy as possible when $n > 1$, since one can benefit from further learning about the model parameters by using the other test items when $\theta$ is not known. Also, he notes in [47] that this classifier is not optimal under a more intuitive loss function, which aims at ensuring that the rate of incorrect labels is minimized for each test item.

(iii) A **Marginalized Predictive Classifier**, denoted by MdPC, assigns the following posterior probability distribution for the label of item $i \in N$

$$p(S_i = c|\mathbf{z}^{(M)}, \mathbf{x}^{(N)}, T) = \sum_{\{S \in \mathcal{S}: S_i = c\}} p(S|\mathbf{z}^{(M)}, \mathbf{x}^{(N)}, T), \quad c = 1, \ldots, k, \tag{3.11}$$

which equals the marginal posterior distribution of the label for the item. This classifier is the optimal rule under the a loss function that sanctions the number of incorrect labels, *i.e.* under loss function of the form

$$L_2(S,\omega) = \sum_{i=1}^{N} I(s_i \neq \omega_i), \tag{3.12}$$

where $s_i$ and $\omega_i$ equal the estimated and true label of the $i$th test item, respectively as noted in[6] [47].

[5]This was first suggested by Geisser, [24] in the context of Gaussian discriminant analysis.

[6]This optimality result was stablished in image analysis by Ripley in [47] wherein it has been shown that a simultaneous labelling of pixels, such that the marginal posterior for each label is maximized, is an optimal classification rule under loss function of the form (3.12). This optimality result, albeit not derived in the predictive classification context, applies also to the classifier MdPC, such that when (3.11) is maximized for all $i \in N$, the classifier minimizes the same expected loss. Also Ripley in [46] considers the related idea of using the remaining test data $\mathbf{x}_1, \ldots, \mathbf{x}_{i-1}, \mathbf{x}_{i+1}, \ldots, \mathbf{x}_n$ to improve learning about model parameters when assigning the label $S_i$ under diagnostic (non-generative) classification. This is distinct from the predictive paradigm considered here and the idea of using marginal posteriors of labels seems not to have appeared earlier in the predictive classification context, except when test items are ordered in time as in speech recognition as [44] and [31]. For the derivation of the same classifier introduced here in the semi-supervised case see [13].

In summary, the differences between the three alternative forms of predictive classifiers can be characterized as follows. Firstly, the MPC is the most basic predictive classifier, which uses only the existing training data to define the posterior distribution of the label of each test item, separately for each test item. In the supervised case this amounts to comparing the posterior predictive probabilities of the observed feature data for each putative class out of the $k$ classes in total. Secondly, a SPC uses the information about the distribution of features shared by all items that are assigned to the same class, *i.e.* data from both the training and all test items simultaneously. When the training data are sparse, the simultaneous predictive probability (3.3) may differ considerably from the product of $n$ marginal predictive probabilities calculated separately according to (3.9) for each test item. The primary reason for this difference is that the simultaneous classifier considers the joint probability of observing any specific combination of feature values over the set of test items that are assigned to the same class, instead of the item-wise marginal probabilities used by the marginal classifier. Finally, the MdPC takes the predictive inferences one step further from the SPC, by considering the labelling decision separately for each test item, such that the labels of the remaining test items are treated as nuisance parameters and hence they should not be treated as fixed when assessing the suitability of a class for an item. Thus the MdPC still acknowledges the predictive modelling of the test data in a simultaneous fashion, but derives from this a decision rule for each individual item by using the marginal posterior distribution of the label. This marginal posterior is in general different from the one resulting in the marginal classifier because the latter ignores the feature data from the remaining items. As discussed above, the optimality of this marginalization strategy follows from the application of an intuitively appealing risk function which aims at minimizing the total number of incorrect labels. The following example will illuminate these arguments.

**Example 3.1.**
Suppose that we have gathered $m = 10$ test items each with $d = 4$ representing features and each feature with $r_j = 3$ different types of categories. Furthermore suppose that there are only $k = 2$ classes. Now we consider the addition of the $n = 3$ test data items to our data collection. While feature values of these items are given, the goal is to find their 'best' representative classes. The summary of the example's information is

$$
\begin{aligned}
&m = 10, \quad n = 3, \quad k = 2, \quad d = 4, \quad |\mathcal{S}| = 8, \\
&\mathcal{X}_j = 1, 2, 3. \quad i = 1, 2, \ldots, 10 \quad j = 1, 2, \ldots 4. \\
&\mathcal{S} = \begin{Bmatrix} S^{(1)} = (1,1,1) & S^{(2)} = (1,1,2) & S^{(3)} = (1,2,1) & S^{(4)} = (1,2,2) \\ S^{(5)} = (2,1,1) & S^{(6)} = (2,1,2) & S^{(7)} = (2,2,1) & S^{(8)} = (2,2,2) \end{Bmatrix},
\end{aligned}
\tag{3.13}
$$

and suppose that our data is given in the format

(3.14)

| | $z_{i1}$ | $z_{i2}$ | $z_{i3}$ | $z_{i4}$ | $T$ |
|---|---|---|---|---|---|
| $\mathbf{z}_1$ | 1 | 2 | 1 | 3 | 1 |
| $\mathbf{z}_2$ | 1 | 2 | 2 | 3 | 1 |
| $\mathbf{z}_3$ | 2 | 1 | 2 | 1 | 1 |
| $\mathbf{z}_4$ | 3 | 3 | 3 | 1 | 1 |
| $\mathbf{z}_5$ | 3 | 3 | 1 | 2 | 1 |
| $\mathbf{z}_6$ | 1 | 3 | 2 | 2 | 2 |
| $\mathbf{z}_7$ | 2 | 3 | 1 | 1 | 2 |
| $\mathbf{z}_8$ | 2 | 2 | 3 | 3 | 2 |
| $\mathbf{z}_9$ | 1 | 2 | 2 | 3 | 2 |
| $\mathbf{z}_{10}$ | 3 | 1 | 2 | 2 | 2 |

(3.15)

| | $x_{i1}$ | $x_{i2}$ | $x_{i3}$ | $x_{i4}$ | $S$ |
|---|---|---|---|---|---|
| $\mathbf{x}_1$ | 1 | 2 | 3 | 1 | $S_1$ |
| $\mathbf{x}_2$ | 1 | 1 | 2 | 1 | $S_2$ |
| $\mathbf{x}_3$ | 2 | 3 | 1 | 1 | $S_3$ |

Now the objective is to identify the optimal value for the classification structure $S = (S_1, S_2, S_3)$. The three classifiers introduced earlier provide three different strategies to solve the problem of finding the optimal value of $S$.

To be more precise, MDC is a step by step optimizer, in a sense that it finds $S_1$, $S_2$, and $S_3$ separately, hence turning out $S$ as a result of these steps. In doing so, it only maximizes the posterior $P(S_i = c|\mathbf{z}^{(M)}, \mathbf{x}_i, T)$ given in (3.9). Here in our case, for finding the $S_2$ for example, it needs only to make a binary comparison between $P(S_2 = 1|\mathbf{z}^{(M)}, \mathbf{x}_2, T)$ and $P(S_2 = 2|\mathbf{z}^{(M)}, \mathbf{x}_2, T)$ and choose the greater posterior probability and return the corresponding class based on this finding.

SPC on the other hand, will provide us with $S$ by a global maximization operation. Based on (3.4) formula, this classifier calculates the posterior probabilities for all the elements $S^{(e)}$ for $e = 1, 2, \ldots, |\mathcal{S}|$, and choose the highest value of these candidates to return the optimal $S$. In our case $P(S = S^{(e)}|\mathbf{x}^{(N)}, \mathbf{z}^{(M)}, T)$ will make an evaluation between 8 elements of $\mathcal{S}$ given in (3.13), where MDC only needs to make 6 comparisons in total.

Lastly, MdPC adheres to an intermediate strategy to solve this problem of classification. It is intermediate in the sense that it finds the labels step by step but in a simultaneously manner. Considering its formula in (3.11), for finding a specific class for a given item, it bases its searching on the entire $\mathcal{S}$, but it will make use of a sum of the posteriori probabilities of all the labellings where a given item is in a particular class. Then it chooses the class that is causing the highest value for that marginal posterior. To illustrate, consider test item number 1, *i.e.*

$\mathbf{x}_1$. For finding the 'best' class for this item with MdPC based on our two potential choices, we have to compare these two entities[7].

$$
\begin{aligned}
\Sigma_1 &= \left\{\begin{array}{ll} & p(S^{(1)}|\mathbf{z}^{(M)},\mathbf{x}^{(N)},T) \\ + & p(S^{(2)}|\mathbf{z}^{(M)},\mathbf{x}^{(N)},T) \\ + & p(S^{(3)}|\mathbf{z}^{(M)},\mathbf{x}^{(N)},T) \\ + & p(S^{(4)}|\mathbf{z}^{(M)},\mathbf{x}^{(N)},T) \end{array}\right\} \\
\Sigma_2 &= \left\{\begin{array}{ll} & p(S^{(5)}|\mathbf{z}^{(M)},\mathbf{x}^{(N)},T) \\ + & p(S^{(6)}|\mathbf{z}^{(M)},\mathbf{x}^{(N)},T) \\ + & p(S^{(7)}|\mathbf{z}^{(M)},\mathbf{x}^{(N)},T) \\ + & p(S^{(8)}|\mathbf{z}^{(M)},\mathbf{x}^{(N)},T) \end{array}\right\}
\end{aligned} \tag{3.16}
$$

If $\Sigma_1 > \Sigma_2$ the choice would be $S_1 = 1$ and in the opposite case, $S_1 = 2$. This is the strategy that should be repeated for the other test items to obtain the probabilistic classification based on this classifier. $\triangle$

### 3.1.2 Predictive distribution under exchangeability

In the previous part we introduced the predictive classifiers without reference to explicit expressions for the actual probabilities as functions of the observed data. Here we derive such expressions for data from multiple finite alphabets. We assume that the observed sequences of features $\mathbf{x}^{(N_c)}$, $\mathbf{z}^{(M_c)}$ assigned to the same class are *unrestrictedly infinitely exchangeable*[8] as defined in [7]. This assumption implies that, if we combine any permutation of the item values $x_{1j},\ldots,x_{n_cj},z_{1j},\ldots,z_{m_cj}$, for a fixed $j = 1,\ldots,d$, with arbitrary corresponding permutations over the remaining features, the same predictive probability mass function for the data is obtained. Furthermore, we also obtain characterizations of the test and training data sets in terms of the sufficient statistics $n_{cjl}, m_{cjl}, l = 1,\ldots,r_j$, where $n_{cjl}$ and $m_{cjl}$ represent the number of copies of value $l$ for feature $j$ observed among the test and training items in the class $c$, respectively.

The unique probabilistic characterization of the data under unrestricted infinite exchangeability assumed to hold over the classes of $S$ then equals by (3.3)

$$
p(\mathbf{x}^{(N)}|\mathbf{z}^{(M)},S,T) = \int_\Theta \prod_{c=1}^{k}\prod_{j=1}^{d}\prod_{l=1}^{r_j} \theta_{cjl}^{n_{cjl}} dF(\theta|\mathbf{z}^{(M)},T), \tag{3.17}
$$

[7]Note that $S^{(1)}, S^{(2)}, S^{(3)}$, and $S^{(4)}$ correspond to all triplets of the $\mathcal{S}$ that their first element is 1 and $S^{(5)}, S^{(6)}, S^{(7)}$, and $S^{(8)}$ on the other hand, represent all those triplets in $\mathcal{S}$ with 2 as their first elements.

[8]For more in-depth discussion about exchangeability see the Chapter 4.

*i.e.,* the test data have a product multinomial[9] likelihood

$$p(\mathbf{x}^{(N)}|\theta, S, T) = \prod_{c=1}^{k}\prod_{j=1}^{d}\prod_{l=1}^{r_j} \theta_{cjl}^{n_{cjl}}, \tag{3.18}$$

where $\theta$ denotes jointly the class-specific multinomial probabilities

$$\theta = \begin{pmatrix} \theta_{c11} & \cdots & \theta_{c1r_1} \\ \vdots & \ddots & \vdots \\ \theta_{cd1} & \cdots & \theta_{cdr_d} \end{pmatrix}, \quad c = 1, \ldots, k, \tag{3.19}$$

and $F(\theta|\mathbf{z}^{(M)}, T)$ is a probability measure over the parameter space $\Theta$. This measure represents our posterior beliefs about the limits of the relative frequencies of observing certain feature values in the different classes, given the observed training data. The extension of the unrestricted exchangeability assumption over the classes implies that the observed values of features within an arbitrary class provide no information about the values observed in any other class. Thus, the assumed exchangeability corresponds to assuming conditional independence of all the features over the considered classes, and leads to a product multinomial likelihood for the data conditional on any $\theta$.

If a class $c$ is void of training data, the corresponding element in the distribution (3.17) is a prior predictive model for the observed items in the class, which is often also termed as the *marginal likelihood* or *evidence.* By using the so called *sufficientness* postulate as [56], the prior beliefs about the relative frequencies for a feature can be explicitly expressed in terms of the Dirichlet($\{\lambda_{cjl}\}_{l=1}^{r_j}$) distribution, which is conjugate for the multinomial likelihood. When this is combined with the product form of (3.17) over features and classes arising from the generalized exchangeability, we can write the predictive probability as

$$p(\mathbf{x}^{(N)}|\mathbf{z}^{(M)}, S, T) = \prod_{c=1}^{k}\prod_{j=1}^{d} \frac{\Gamma(m_c + 1)}{\Gamma(n_c + m_c + 1)} \prod_{l=1}^{r_j} \frac{\Gamma(n_{cjl} + m_{cjl} + \lambda_{cjl})}{\Gamma(m_{cjl} + \lambda_{cjl})}, \tag{3.20}$$

where it is assumed for simplicity that the Dirichlet hyperparameters are constants defined as $\lambda_{cjl} = 1/r_j$. We note that under MPC the predictive probability is a special case of (3.20) with $n = 1, \mathbf{x}^{(N)}$ consisting thus of a single data vector only, we also urge reader to compare this formula with what is discussed in second chapter as Beta-Binomial distribution (2.14).

It is also necessary to consider the prior distribution of $S$. For this we consider the simple case that labels are independent of each other. In this case a prior symmetry of labels with respect to their ordering can be represented by setting the joint probability of $S$ equal to

$$p(S|\phi) = \prod_{c=1}^{k} \phi_c^{n_c}, \tag{3.21}$$

where $n_c$ is the count of labels $\sum_{t=1}^{n} I(s_t = c)$, $\phi_c$ represents the limiting relative frequency of observing an item with label $c$, $\phi = (\phi_1, \dots, \phi_c), \phi \in \mathbf{\Phi}$ and $\sum_{c=1}^{k} \phi_c = 1$. By introducing a Dirichlet density to represent the uncertainty about $\phi$ according to

$$Dir(\phi|\beta_1, \dots, \beta_k) = \frac{\Gamma(\sum_{c=1}^{k} \beta_c)}{\prod_{c=1}^{k} \Gamma(\beta_c)} \prod_{c=1}^{k} \phi_c^{\beta_c - 1}, \tag{3.22}$$

with $\beta_c > 0$, the prior distribution of $S$ equals

$$\begin{aligned} p(S|T) &= \int_{\phi \in \mathbf{\Phi}} p(S|\phi) p(\phi|T) d\phi \\ &= \frac{\Gamma(m + \sum_{c=1}^{k} \beta_c)}{\Gamma(n + m + \sum_{c=1}^{k} \beta_c)} \prod_{c=1}^{k} \frac{\Gamma(n_c + m_c + \beta_c)}{\Gamma(m_c + \beta_c)}, \end{aligned} \tag{3.23}$$

where $m_c$ is defined analogously to $n_c$.

[9]For more information about common probability distributions see Appendix C.

At this point we can revisit example 3.1 and with assumption given in this section, find the classification structure $S$.

**Example 3.2.** Consider the setting given in (3.13), (3.14), and (3.15). Now we want to find the 'best' $S = (S_1, S_2, S_3)$ for our three test data. We consider the same matrix format for the frequency of observations as parameters given in (3.19) as

$$m^{(c)} = \begin{pmatrix} m_{c11} & \cdots & m_{c1r_1} \\ \vdots & \ddots & \vdots \\ m_{cd1} & \cdots & m_{cdr_d} \end{pmatrix}, \quad n^{(c)} = \begin{pmatrix} n_{c11} & \cdots & n_{c1r_1} \\ \vdots & \ddots & \vdots \\ n_{cd1} & \cdots & n_{cdr_d} \end{pmatrix}, \quad c = 1, 2 \tag{3.24}$$

that $m^{(c)}$ and $n^{(c)}$ contain the elements $m_{cjl}$ and $n_{cjl}$ defined in this section, respectively. Based on the training data, $m^{(c)}$ would essentially be

$$m^{(1)} = \begin{pmatrix} 2 & 1 & 2 \\ 1 & 2 & 2 \\ 2 & 2 & 1 \\ 2 & 1 & 2 \end{pmatrix}, \quad m^{(2)} = \begin{pmatrix} 2 & 2 & 1 \\ 1 & 2 & 2 \\ 1 & 3 & 1 \\ 1 & 2 & 2 \end{pmatrix}. \tag{3.25}$$

This is obvious that based on different elements of $\mathcal{S}$ we would have two distinct $n^{(1)}$ and $n^{(2)}$ matrices.

(3.26)

$$n^{(1)}_{S^{(1)}} = \begin{pmatrix} 2 & 1 & 0 \\ 1 & 1 & 1 \\ 1 & 1 & 1 \\ 3 & 0 & 0 \end{pmatrix}, n^{(2)}_{S^{(1)}} = \begin{pmatrix} 0 & 0 & 0 \\ 0 & 0 & 0 \\ 0 & 0 & 0 \\ 0 & 0 & 0 \end{pmatrix}, n^{(1)}_{S^{(2)}} = \begin{pmatrix} 2 & 0 & 0 \\ 1 & 1 & 0 \\ 0 & 1 & 1 \\ 2 & 0 & 0 \end{pmatrix}, n^{(2)}_{S^{(2)}} = \begin{pmatrix} 0 & 1 & 0 \\ 0 & 0 & 1 \\ 1 & 0 & 0 \\ 1 & 0 & 0 \end{pmatrix}$$

$$n^{(1)}_{S^{(3)}} = \begin{pmatrix} 1 & 1 & 0 \\ 0 & 1 & 1 \\ 1 & 0 & 1 \\ 2 & 0 & 0 \end{pmatrix}, n^{(2)}_{S^{(3)}} = \begin{pmatrix} 1 & 0 & 0 \\ 1 & 0 & 0 \\ 0 & 1 & 0 \\ 1 & 0 & 0 \end{pmatrix}, n^{(1)}_{S^{(4)}} = \begin{pmatrix} 1 & 0 & 0 \\ 0 & 1 & 0 \\ 0 & 0 & 1 \\ 1 & 0 & 0 \end{pmatrix}, n^{(2)}_{S^{(4)}} = \begin{pmatrix} 1 & 1 & 0 \\ 1 & 0 & 1 \\ 1 & 1 & 0 \\ 2 & 0 & 0 \end{pmatrix},$$

$$n^{(2)}_{S^{(8)}} = \begin{pmatrix} 2 & 1 & 0 \\ 1 & 1 & 1 \\ 1 & 1 & 1 \\ 3 & 0 & 0 \end{pmatrix}, n^{(1)}_{S^{(8)}} = \begin{pmatrix} 0 & 0 & 0 \\ 0 & 0 & 0 \\ 0 & 0 & 0 \\ 0 & 0 & 0 \end{pmatrix}, n^{(2)}_{S^{(7)}} = \begin{pmatrix} 2 & 0 & 0 \\ 1 & 1 & 0 \\ 0 & 1 & 1 \\ 2 & 0 & 0 \end{pmatrix}, n^{(1)}_{S^{(7)}} = \begin{pmatrix} 0 & 1 & 0 \\ 0 & 0 & 1 \\ 1 & 0 & 0 \\ 1 & 0 & 0 \end{pmatrix}$$

$$n^{(2)}_{S^{(6)}} = \begin{pmatrix} 1 & 1 & 0 \\ 0 & 1 & 1 \\ 1 & 0 & 1 \\ 2 & 0 & 0 \end{pmatrix}, n^{(1)}_{S^{(6)}} = \begin{pmatrix} 1 & 0 & 0 \\ 1 & 0 & 0 \\ 0 & 1 & 0 \\ 1 & 0 & 0 \end{pmatrix}, n^{(2)}_{S^{(5)}} = \begin{pmatrix} 1 & 0 & 0 \\ 0 & 1 & 0 \\ 0 & 0 & 1 \\ 1 & 0 & 0 \end{pmatrix}, n^{(1)}_{S^{(5)}} = \begin{pmatrix} 1 & 1 & 0 \\ 1 & 0 & 1 \\ 1 & 1 & 0 \\ 2 & 0 & 0 \end{pmatrix},$$

To classify the test item simultaneously, *i.e.* based on SPC, with referring to (3.4), we will see that we need to calculate two components for each element of the $\mathcal{S}$. These components are the prior of labels $p(S|T)$ and predictive posterior of the test data $p(\mathbf{x}^{(N)}|\mathbf{z}^{(M)}, S, T)$. First, consider $p(S|T)$.

Since in our case there are only two classes, the prior given in (3.22) is a Beta distribution and for convenience we set its (hyper) parameters to 1 that further simplify this prior to standard Uniform distribution. Based on this choice, the formula given in (3.23) will be

$$p(S|T) = \frac{\Gamma(m+\alpha+\beta)}{\Gamma(n+m+\alpha+\beta)} \frac{\Gamma(n_1+m_1+\alpha)}{\Gamma(m_1+\alpha)} \frac{\Gamma(n_2+m_2+\beta)}{\Gamma(m_2+\beta)}, \tag{3.27}$$

that $\alpha$ and $\beta$ are the hyperparameters of the prior $P(\phi)$ which are set to 1. In our case we can further simplify this predictive prior distribution as

$$p(S|T) = \frac{\Gamma(12)}{\Gamma(15)} \frac{(n_1+6)}{\Gamma(6)} \frac{\Gamma(n_2+6)}{\Gamma(6)},$$

that as it shows, it is a function of the frequency vector or the test data[10]. We can calculate now the prior probabilities of the simultaneous structures $p(S|T)$ for all

[10]Note that the likelihood of labels $T$ are considered as unordered multinomial distribution that is used in the (3.23) formula for calculating the posterior of the $\phi$.

the different elements of $\mathcal{S}$ based on the information given in matrices of (3.26). For example

$$p(S = S^{(3)}|T) = p(S = (1,2,1)|T) = \frac{\Gamma(12)}{\Gamma(15)}\frac{\Gamma(8)}{\Gamma(6)}\frac{\Gamma(7)}{\Gamma(6)} = \frac{3}{26},$$

and this is because for $S = S^{(3)}$ we have $n_1 = 2$ and $n_2 = 1$. The list of all the values of $p(S|T)$, calculated in the same manner is given in table (3.29).

Secondly, we need to calculate the values of the predictive posterior of the test data as mentioned earlier. In doing so, we refer to the formula (3.20) and note further that here we consider $\lambda_{cjl} = 1$ as the hyperparameters of our Dirichlet prior. Hence, our function will essentially be

$$(3.28) \qquad p(\mathbf{x}^{(N)}|\mathbf{z}^{(M)}, S, T) = \prod_{c=1}^{2}\prod_{j=1}^{4}\frac{\Gamma(m_c+3)}{\Gamma(n_c+m_c+3)}\prod_{l=1}^{3}\frac{\Gamma(n_{cjl}+m_{cjl}+1)}{\Gamma(m_{cjl}+1)}.$$

For calculating this quantity for every distinct elements of $\mathcal{S}$ we should now refer to matrices given in (3.25) and (3.26). For example

$$p(\mathbf{x}^{(N)}|\mathbf{z}^{(M)}, S = S^{(3)}, T) = (1.695421)^{-6} =$$
$$(\frac{\Gamma(8)}{\Gamma(10)})^4\Big(\frac{3!2!2!}{2!1!2!}\frac{1!3!3!}{1!2!2!}\frac{3!2!2!}{2!2!1!}\frac{4!1!2!}{2!1!2!}\Big)(\frac{\Gamma(8)}{\Gamma(9)})^4\Big(\frac{3!2!1!}{2!2!1!}\frac{2!2!2!}{1!2!2!}\frac{1!4!1!}{1!3!1!}\frac{2!2!2!}{1!2!2!}\Big).$$

The table (3.29) lists the different values of this quantity as a function of elements of the $\mathcal{S}$ and also provide the multiplications of the two columns as the numerator of the (3.4). Note that it is not necessary to calculate the denominator $\sum_{S\in\mathcal{S}} p(\mathbf{x}^{(N)}|\mathbf{z}^{(M)}, S, T)p(S|T)$, since our task here is to maximize the function and this value will appear as a multiplication constant for all posterior probabilities.

(3.29)

| $\mathbf{S^{(e)}}$ | $\mathbf{a} = p(\mathbf{x}^{(N)}\|\mathbf{z}^{(M)}, \mathbf{S^{(e)}}, T)$ | $\mathbf{b} = p(\mathbf{S^{(e)}}\|T)$ | $-\log(\mathbf{a} * \mathbf{b})$ | $\mathbf{a} * \mathbf{b}$ |
|---|---|---|---|---|
| $S^{(1)}$ | $7.233796 \times 10^{-8}$ | 0.1538462 | 18.31372 | $1.112892 \times 10^{-8}$ |
| $S^{(2)}$ | $9.889956 \times 10^{-7}$ | 0.1153846 | 15.98606 | $1.141149 \times 10^{-7}$ |
| $S^{(3)}$ | $1.695421 \times 10^{-6}$ | 0.1153846 | 15.4470* | $1.956255 \times 10^{-7}$ |
| $S^{(4)}$ | $1.271566 \times 10^{-6}$ | 0.1153846 | 15.73475 | $1.467191 \times 10^{-7}$ |
| $S^{(5)}$ | $3.532127 \times 10^{-8}$ | 0.1153846 | 19.31826 | $4.075531 \times 10^{-9}$ |
| $S^{(6)}$ | $9.536743 \times 10^{-7}$ | 0.1153846 | 16.02243 | $1.100393 \times 10^{-7}$ |
| $S^{(7)}$ | $1.695421 \times 10^{-6}$ | 0.1153846 | 15.4470* | $1.956255 \times 10^{-7}$ |
| $S^{(8)}$ | $9.259259 \times 10^{-7}$ | 0.1538462 | 15.76427 | $1.424501 \times 10^{-7}$ |

Based on these result we choose either of classification structures $S^{(3)}$ and $S^{(7)}$ with equal probability as they turned to give the highest joint posterior probabilities as marked by asterisk in the table. So our 'best' guess for the labels of the test data $\mathbf{x}^{(3)}$ based on SPC is either cases (1,2,1) or (2,2,1).

Based on this table we can even obtain the 'best' prediction for $S$ based on MdPC. Referring to its formula given in (3.11), we see that the components of the sum on the left side of the equation are the normalized values of the $\mathbf{a}*\mathbf{b}$ column of the above table where the normalizing constant is $\sum_{S\in\mathcal{S}} p(\mathbf{x}^{(N)}|\mathbf{z}^{(M)}, S, T)p(S|T)$. Again, its computation is not necessary, since we have to make a comparison between the summation of posterior probabilities for two candidates where a multiplicative constant in each of the sum terms is of no consequence. In structure $S = (S_1, S_2, S_3)$, let us first find $S_1$, but before that, we provide the structure set $\mathcal{S}$ once again for easier reference.

$$\mathcal{S} = \begin{Bmatrix} S^{(1)} = (1,1,1) & S^{(2)} = (1,1,2) & S^{(3)} = (1,2,1) & S^{(4)} = (1,2,2) \\ S^{(5)} = (2,1,1) & S^{(6)} = (2,1,2) & S^{(7)} = (2,2,1) & S^{(8)} = (2,2,2) \end{Bmatrix}.$$

With what is noted in (3.16) and its following argument, we first form the unnormalized $\Sigma_1^*$ and $\Sigma_2^*$ and further refer to these values for each of the three test items as superscript (1), (2), and (3), respectively. Based on the table values we can derive the comparative values as

$$\begin{array}{lll} \Sigma_1^{*(1)} = 4.675884 \times 10^{-7} & \Sigma_1^{*(2)} = 2.393587 \times 10^{-7} & \Sigma_1^{*(3)} = 4.064555 \times 10^{-7} \\ \Sigma_2^{*(1)} = 4.521904 \times 10^{-7} & \Sigma_2^{*(2)} = 6.804202 \times 10^{-7} & \Sigma_2^{*(3)} = 5.133234 \times 10^{-7} \end{array}$$

Based on these findings we conclude that the best structure based on MdPC is $S = (1, 2, 2)$. △

## 3.2 Asymptotic Properties of Predictive Classifiers

### 3.2.1 Infinite training data

Here we characterize the behaviour of the simultaneous predictive classifier in relation to the marginal classifiers when the amount of the training data increases. Recall that the logarithm of the data predictive (3.20) for the simultaneous clas-sifier SPC equals under the previous derivations

$$(3.30)\quad \log p(\mathbf{x}^{(N)}|\mathbf{z}^{(M)}, S, T) = \sum_{c=1}^{k}\sum_{j=1}^{d} \log \frac{\Gamma(m_c+1)}{\Gamma(n_c+m_c+1)} + \sum_{c=1}^{k}\sum_{j=1}^{d}\sum_{l=1}^{r_j} \log \frac{\Gamma(n_{cjl}+m_{cjl}+\lambda_{cjl})}{\Gamma(m_{cjl}+\lambda_{cjl})}.$$

We can derive a comparable expression for the logarithm of the data predictive probability (3.20) of the marginal classifier MPC by considering

$$(3.31)\quad \log p(\mathbf{x}_1,\ldots,\mathbf{x}_n|\mathbf{z}^{(M)}, S, T) = \sum_{c=1}^{k}\sum_{i:S_i=c} \log p(\mathbf{x}_i|\mathbf{z}^{(M)}, T, S_i = c),$$

where $S$ now refers to the classification structure resulting from the assignment of the $n$ items into the classes one by one. Using Stirling's approximation[11] to the gamma function terms in (3.30) and (3.31), it follows that for any particular classification structure $S$, the difference in the logarithms of the data predictive probability between the simultaneous and marginal classifiers tends to zero as the amount of training data increases, *i.e.*

$$\lim_{m\to\infty} \log p(\mathbf{x}^{(N)}|\mathbf{z}^{(M)}, S, T) - \log p(\mathbf{x}_1,\ldots,\mathbf{x}_n|\mathbf{z}^{(M)}, S, T) = 0,$$

when the sampling process of training data from each class is infinitely unrestrictedly exchangeable. The limit of the relative frequencies of feature values exist and we assume that they are strictly positive. Consequently, the classifier MPC and SPC will become equivalent with increasing training data set size, as the optimal classification according to one will coincide with that of the other, which is explicitly given in [13] as a theorem which we restate here.

**Theorem 1.** *Asymptotic equivalence of supervised predictive classifiers. Assume that all limits of the relative frequencies of feature values are strictly positive under an infinitely exchangeable sampling process*

$$\frac{m_{cjl}}{m_c} \underset{m\to\infty}{\to} \beta_{cjl} > 0, \quad c = 1,\ldots,k, \quad j = 1,\ldots,d, \quad l = 1,\ldots,r_j.$$

*Then, the optimal classification under SPC and MPD are asymptotically equals as the amount of training data $m$ tends to infinity.*

*Proof.* The proof is given in the Appendix A. □

[11]For its mathematical definition see Appendix B.

The limiting form of the predictive classifier coincides with the classifier obtained assuming *i.i.d.* sampling under fixed probability measure. Intuitively, the above theorem tells us that no additional inductive learning can take place in the optimal simultaneous classifier when the amount of training information increases sufficiently. This has a particular consequence in classification applications as numerical implementation of simultaneous classifier rules is considerably more computation intensive than of marginal classifier. However, the expected accuracies of the two classifiers may differ when the amount of training data is small. For extensive investigation of the semi-supervised case and learning algorithms for both cases see [13].

### 3.2.2 Infinite test data

In the last section we showed that MPC and SPC for non-sequential data become equivalent rules when the amount of training data $m$ tends to infinity under an infinitely exchangeable sampling process. Such a limiting behaviour indicates that no considerable additional information can be obtained from observed features of test items in the presence of a sufficient amount of training information. Here we generalize that result to establish the behaviour of sequential predictive classifiers when the amount of test data increases, instead of the amount of training data, which remains here fixed and bounded.

Consider a sequence of $n$ items with the feature vectors $\mathbf{x}^{(N)} = \{\mathbf{x}_1, \ldots, \mathbf{x}_n\}$, and a classification assignment $S = \{s_1, \ldots, s_n\}$. For the subsequent data with $\delta$ items, the feature vectors $\{\mathbf{x}_{n+1}, \mathbf{x}_{n+2}, \ldots, \mathbf{x}_{n+\delta}\}$ and a label sequence $\{s_{n+1}, \ldots, s_{n+\delta}\}$ are denoted by $\mathbf{x}^{(\delta)}$ and $s^{(\delta)}$, respectively. Behaviour of the SPC based on (3.23) and (3.20) is explicitly characterized in the theorem and its corollary below as they appear in [12].

**Theorem 2.** *Assume that all limits of relative frequencies of feature values are strictly positive under an infinitely exchangeable sampling process and that the prior predictive distribution of labels equals (3.23)*

$$\frac{n_{cjl}}{n_c} \to \eta_{cjl} > 0, \quad c = 1, \ldots, k, \quad j = 1, \ldots, d, \quad l = 1, \ldots, \chi_j. \tag{3.32}$$

*Then,*

$$p(s^{(\delta)}|\mathbf{x}^{(N)}, \mathbf{x}^{(\delta)}, \mathbf{z}^{(M)}, S, T) \underset{n\to\infty}{\to} \prod_{t=n+1}^{n+\delta} p(s_t|\mathbf{x}^{(N)}, \mathbf{x}_t, \mathbf{z}^{(M)}, S, T),$$

*and for any* $t \in (n, n+\delta] \cap \mathbb{Z}$

$$p(s_t|\mathbf{x}^{(N)}, \mathbf{x}^{(\delta)}, \mathbf{z}^{(M)}, S, s^{(\delta)\setminus\{t\}}, T) \underset{n\to\infty}{\to} p(s_t|\mathbf{x}^{(N)}, \mathbf{x}_t, \mathbf{z}^{(M)}, S, T),$$

*where it is assumed that the label sequence $S$ satisfies the condition that $n \to \infty$ implies $n_c \to \infty, c = 1, ..., k$.*

*Proof.* The proof is given in the Appendix A. □

The two limit results thus establish that given a sufficiently long history of test items, SPC of subsequent data will not enable additional inductive learning for the classifier. For the predictive distribution of the MdPC, we have the following corollary of the above theorem that is denoting the equivalence of SPC and MdPC in the scenario depicted in the latter theorem.

**Corollary 1**. *Under the same assumptions as in Theorem 2, SPC and MdPC are equivalent for any $t \in (n, n + \delta] \cap Z$.*

*Proof.* The proof is given in the Appendix A. □

# Chapter 4

# Partition Exchangeability

A major difficulty for currently existing theories of inductive inference involves the question of what to do when novel, unknown, or previously unsuspected phenomena occur. In this chapter one particular instance of this difficulty that has the highest degree of connection with the type of prediction that is considered in the consecutive chapter, is referred to as sampling of species problem in [53].

The classical probabilistic theories of inductive inference due to Laplace, Johnson, De Finetti, and Carnap adopt a model of simple enumerative induction in which there is a prespecified number of types of species which may be observed. But, realistically, this is often not the case. In 1838 the English mathematician Augustus De Morgan proposed a modification of the Laplacian model to accommodate situations where the possible types of species to be observed are not assumed to be known in advance; but he did not advance a justification for his solution. In this chapter a general philosophical approach to such problems is discussed, drawing on work of the mathematician J. F. C. Kingman. It then emerges that the solution advanced by De Morgan has a very deep justification. The key idea is that although 'exchangeable' random *sequences* are the right objects to consider when all possible outcome-types are known in advance, exchangeable random *partitions* are the right objects to consider when they are not. The classical theory has several basic elements: a representation theorem for the general exchangeable sequence (the de Finetti representation theorem), a distinguished class of sequences, and a corresponding rule of succession. The new theory has parallel basic elements: a representation theorem for the general exchangeable random partition (the Kingman representation theorem), a distinguished class of random partitions, and a rule of succession which corresponds to De Morgan's rule. In this chapter we consider both of these cases in a way to equip ourselves with necessary tools so that we can formulate a way of tackling predictive classification when all the possible values of the discrete features are not known *a priori*, that is the topic of the next chapter. Although in previous chapter the knowledge of the de Finetti type of exchangeability is assumed for the reader, here we investigate this type of exchangeability in more detail and also as a tool to empower us with a parallel line of metaphor and comparison with the partition exchangeability. We further note that this chapter is mainly based on the content of the tutorial work of Zabell in [53]. For more in depth discussion on the subject see [53] and [38-41]. In the first section we present the work of De Morgan and Laplace about the sampling from an urn plus the De Morgan process that towards the end of the chapter will be shown to be more than an arbitrary derivation. The second section briefly reviews the classical probabilistic account of induction for a fixed number of categories. The third section will discuss the partition exchangeability while the Kingman representation theorem and the Ewens sampling formula is presented in the last section.

## 4.1 The De Morgan Process

We start with *Laplace's rule of succession* that is prompted as the answer to the question of resemblance of the future to past. In its simple form it states that if an event has occurred $n$ times out of $N$ experiences in the past, then the probability that it will occur the next time is $(n+1)/(N+2)$, implying the dichotomousness of the outcomes. A more complex form of the rule assumes that in the $N$ experiments done so far, we have observed $d$ different classes $c_1, \dots, c_d$, each with $\mathbf{n} = (n_1, \dots, n_d)$ frequencies. Then the probability that an outcome of the $j$th type will occur on the next trial is

$$\text{(4.1)} \qquad \textbf{Laplace's Rule} \quad p(X_{(N+1)} = c_j | \mathbf{n}) = \frac{n_j + 1}{N + d}.$$

Laplace considered the multinomial case where there are three or more categories, but his analysis is limited to those instances where the number of categories is fixed in advance. De Morgan in [18] with referring to (4.1) noted "There remains, however, an important case not yet considered; suppose that having obtained $d$ sorts in $N$ drawings, and $d$ sorts only, we do not yet take it for granted that these are all the possible cases, but allow ourselves to imagine there may be sorts not yet come out." Thus in contrast, he proposed a simple way of dealing with the possibility of an unknown number of categories in [17] as

$$\text{(4.2)} \qquad \textbf{De Morgan's Rule} \quad p(X_{(N+1)} = c_j | \mathbf{n}) = \frac{n_j + 1}{N + d + 1}.$$

That is one creates an additional category : 'new species not yet observed', which has a probability of $1/(N+d+1)$ of occurring. Note that (4.2) implies $d$ as the known number of classes prior to sampling the next observation.

### The De Morgan process

The urn model denoting the sequence of probabilities suggested by De Morgan is as follow. Consider an urn with one black ball (the 'mutator'), and $d$ additional balls, each of a different color, say, $c_1, \ldots, c_d$. We reach into the urn, pick a ball at random, and return it to the urn *together with a new ball*, according to the following rule:

- if a coloured ball is drawn, then it is replaced together with another of the *same* color.
- If the mutator is drawn, then it is replaced together with another ball of an *entirely* new color.

We can have a practical representation for this model. One can consider the coloured balls correspond to species known to exist so far; selecting a ball of a given color corresponds to observing the species represented by that color; selecting the mutator corresponds to observing a hitherto unknown species. Clearly this sequence of operations generates the probabilities De Morgan suggests. After $N$ drawings, there are $N + d + 1$ balls in the urn, because we started out with $d$ (the coloured balls) $+1$ (the mutator) and have added $N$ since. Because we are choosing balls at random, each has a probability of 1 in $N + d + 1$ of being selected. The number of colors is gradually changing, but if there are $n_j + 1$ balls of a specific type, then the probability of observing that type at the next draw is the one given by De Morgan. On the other hand, since there is always only one mutator, the probability of it being selected (the probability that a new species is observed) is $1/(N+d+1)$. This of course is adjustable to any desirable probability $\theta/(N+d+\theta)$ by assigning appropriate value of $\theta$. This process generates the probabilities specified by De Morgan, so we shall call it the *De Morgan process*. Thus, De Morgan's description is consistent and it will further shown that it is a very special process possessing many attractive features that separate it from many other hypothetical processes that one can invent from a broad spectrum of processes for such a problem.

## 4.2 Exchangeable Random Sequences

Efforts to describe enumerative induction in probabilistic terms go back to Bayes and Laplace, but the attempt was more fruitful at the hands of the twentieth-century Italian mathematician and philosopher Bruno de Finetti. De Finetti's insight was that those situations in which the simplest forms of enumerative induction are appropriate are captured by the mathematical concept of 'exchangeability', and that the mathematical structure of such sequences is readily described.

### 4.2.1 The de Finetti representation theorem

Let $X_1, \ldots, X_N, \ldots$ be an infinite sequence of random variables taking on any of a finite number of values, say[1] $c_1, \ldots, c_d$. The sequence is said to be *exchangeable* if for every $N$ the 'cylinder set' probabilities $p(X_1 = e_1, \ldots, X_N = e_N) = p(e_1, \ldots, e_N)$ are invariant under all possible permutation of the time index. Put another way, two sequences have the same probability if one is a *rearrangement* of the other. If the outcomes are thought of as letters in an alphabet, then this means that all words of the same length having the same letters have the same probability. Given a sequence of possible outcomes $e_1, \ldots, e_N$, let $n_j$ denote the number of times the $j$th type occurs in the sequence. The frequency vector $\mathbf{n} = (n_1, \ldots, n_d)$ plays a key role in exchangeability. First, it provides an equivalent characterization of exchangeability, since given any two sequences, say $\mathbf{e} = (e_1, \ldots, e_N)$ and $\mathbf{e}^* = (e_1^*, \ldots, e_N^*)$, one can be obtained from the other by rearrangement if and only if the two have the same frequency vector. Thus, $p$ is exchangeable if and only if two sequences having the same frequency vector have the same probability. This gives us the following formal definition.

**Definition 3.** *A probability function $p$ is* ***exchangeable*** *if for every $N$ the cylinder set probabilities $p(X_1 = e_1, \ldots, X_N = e_N)$ are invariant under permutations of the* ***time*** *index.*

This definition underlies the fact that exchangeability is bound with probability measure defined for a sequence of random variables. In the language of theoretical statistics, the observed frequency counts $n_j = n_j(X_1, \ldots, X_N)$ are *sufficient statistics* for the sequence $\{X_1, \ldots, X_N\}$, in the sense that probabilities conditional on the frequency counts depend only on $\mathbf{n}$, and are independent of the choice of exchangeable $p$: given a particular value of the frequency vector, the only sequences possible are those having this frequency vector, and each of these, by exchangeability, is assumed equally likely. The number of such sequences is given by the *multinomial coefficient* $N!/(n_1!n_2! \ldots, n_d!)$; and, thus the probability of such a sequence is

$$p(X_1, \ldots, X_N|\mathbf{n}) = \frac{n_1!n_2! \ldots n_d!}{N!}.$$

The structure of exchangeable sequences is actually quite simple. Let

$$\Delta_d =: \{(p_1, \ldots, p_d) : p_i \geq 0 \quad \& \quad p_1 + \cdots + p_d = 1\},$$

[1]These are the possible *categories* or *cells* into which the outcomes of the sequence are classified, and might denote different species in an ecosystem, or words in a language.

denote the $d$-simplex of probabilities on $d$ elements[2]. Every element of the simplex determines a multinomial probability, and the general exchangeable probability is a mixture of these. This is the content of a theorem due to de Finetti; such that if an infinite sequence of $d$-valued random variables $X_1, X_2, \dots$ is exchangeable, and $(n_1, \dots, n_d)$ is the vector of frequencies for $\{X_1, \dots, X_N\}$, then the infinite limiting frequency

$$Z =: \lim_{N\to\infty} \left(\frac{n_1}{N}, \dots, \frac{n_d}{N}\right),$$

exists almost surely; and, if $F(A) = p(Z \in A)$ denotes the distribution of this limiting frequency, then

$$p(X_1 = e_1, \dots, X_N = e_N) = \int_{\Delta_d} p_1^{n_1} p_2^{n_2} \dots p_d^{n_d} dF(p_1, p_2, \dots, p_{d-1}).$$

The use of such integral representations of course predates de Finetti; de Finetti's contribution was to give a philosophical justification for their use, based on the concept of exchangeability, one not appealing to objective chances or second-order probabilities to explain the nature of the multinomial probabilities appearing in the mixture (see e.g., [54] and [55]).

### 4.2.2 The rule of succession

Once the prior measure $dF$ has been implicitly of explicitly specified[3], one can immediately calculate the *predictive probabilities* that it gives rise to

$$p(X_{N+1} = c_i | X_1, X_2, \dots, X_N) = p(X_{N+1} = c_i | \mathbf{n}).$$

Such a conditional probability is sometimes called a 'rule of succession'. For example, in the case of Bayes-Laplace prior ($d = 2$), a simple integration immediately yields Laplace's rule of succession.

$$p(X_{N+1} = c_1 | \mathbf{n}) = \frac{n_1 + 1}{N + 2},$$

and in case of considering flat prior due to Laplace for $d \geq 3$ we have

$$p(X_{N+1} = c_i | \mathbf{n}) = \frac{n_i + 1}{N + d}.$$

[2]Note here the slight change in notation as $d$ did earlier refer to the number of features in our data vectors.

[3]The earliest and best-known prior is the so-called 'Bayes-Laplace prior', which assumes that there are two categories, say 'success' and 'failure' (so that $d$ =2), and takes $dF(p) = dp$. The generalization of this prior to $d \geq 3$ (*i.e.* 'flat' prior $dF(p_1, \dots, p_d) = dp_1 \dots dp_d$) is first introduced by Laplace in 1778 (see e.g. [54] and [53] for more discussion about prior).

Further one can consider the less stringent postulate as Johnson did in [36] to get the 'sufficient' postulate as

$$p(X_{N+1} = c_i|\mathbf{n}) = f(n_i, N),$$

which is stating that the only relevant information conveyed by the sample for predicting whether the next outcome will fall into a given category $c_i$, is the number of outcomes observed in that category till now; and any knowledge of how outcomes not in that category distribute themselves among the reminder is irrelevant. As a consequence of sufficient postulate, Johnson was able to derive, the corresponding rule of succession: if $X_1, X_2, \ldots$ is an exchangeable sequence satisfying the sufficientness postulate, and $d \geq 3$, then (assuming that all cylinder set probabilities are positive so that the relevant conditional probabilities exist)

$$p(X_{N+1} = c_i|\mathbf{n}) + \frac{n_i + \alpha}{N + d\alpha}$$

The corresponding measure in the de Finetti representation in this case is the *symmetric Dirichlet*[4] *prior* with parameter $\alpha$.

## 4.3 Partition Exchangeability

Johnson's sufficientness postulate attempts to capture the concept of prior ignorance about categories. Despite its attractiveness, however, it is far from clear that Johnson's sufficientness postulate is a necessary condition for such a state of ignorance. Is it possible to further weaken the notion of absence of information about the categories? A natural idea is that ignorance about *categories* should result in a symmetry of beliefs similar to that captured by de Finetti's notion of exchangeability with respect to *times*. This suggests the following definition.

**Definition 4.** *A probability function $p$ is* ***partition exchangeable*** *if for every $N$ the cylinder set probabilities $p(X_1 = e_1, \ldots, X_N = e_N)$ are invariant under permutations of the* ***time*** *and* ***category*** *index.*

For example, if we are rolling a die (so that $d = 6$), and our subjective probabilities for the various outcomes are partition exchangeable, then

$$p(6, 4, 6, 4, 4, 5, 1, 2, 5) = p(1, 1, 1, 2, 2, 3, 3, 4, 5).$$

This can be seen by first arranging the sequence

$$\{6, 4, 6, 4, 4, 5, 1, 2, 5\},$$

[4]See Appendix C.

into 'regular position'

$$\{4, 4, 4, 5, 5, 6, 6, 1, 2\},$$

(*i.e.*, descending order of observed frequency for each face); and then follow this up by the category permutation

$$1 \to 4 \to 1, \quad 2 \to 5 \to 2, \quad 3 \to 6 \to 3,$$

which can be more compactly written as $(1, 4), (2, 5), (3, 6)$.

The 'sufficient statistic' for a partition exchangeable sequence are the *frequencies of the frequencies* or *abundances*,

$$\mathbf{a_r} =: \text{number of } \mathbf{n_j} \text{ equal to } \mathbf{r}.$$

**Example 4.1.** Suppose one observes the sequence $5, 2, 6, 1, 2, 3, 5, 1, 1, 2$. Then:

$N = 10; d = 6.$
$n_1 = 3, n_2 = 3, n_3 = 1, n_4 = 0, n_5 = 2, n_6 = 1.$
$\mathbf{n} = (3, 3, 1, 0, 2, 1)\text{``} = \text{''}0^1 1^2 2^1 3^2.$
$a_0 = 1, a_1 = 2, a_2 = 1, a_3 = 2, \ldots, a_{10} = 0.$
$\mathbf{a} = (1, 2, 1, 2, 0, \ldots, 0)$

△

We call the **a**-vector the *partition vector.* This partition vector **a** plays the same role relative to partition exchangeable sequence that the frequency vector **n** plays for ordinary exchangeable sequences; that is, two sequences are equivalent, in the sense that one can be obtained from the other by a permutation of the time set and a permutation of the category set, if and only if the two sequences have the same partition vector. Thus, an alternative characterization of partition exchangeability is that: *all sequences having the same partitions vector have the same probability.* The frequencies of frequencies, furthermore, are the sufficient statistics for a partition exchangeable sequence, since probabilities conditional on the partition vector $\mathbf{a} = (a_0, a_1, \ldots, a_d)$ are independent of $p$. Finally, given a partition vector **a**, the only possible sequences have **a** as partition vector and each of these is equally likely.

According to the de Finetti representation theorem, a partition exchangeable sequence, being exchangeable can be represented by a mixing measure $dF$ on the $d$-simplex $\Delta_d$. An important subset of the $d$-simplex in the partition exchangeable case is the subsimplex of ordered probabilities:

$$\Delta_d^* =: \{(p_1^*, \ldots, p_d^*) : p_1^* \geq \cdots \geq p_d^* \geq 0 \quad \& \quad \sum_j p_j^* = 1\},$$

In the partition exchangeable case, once the prior $dF$ is known on the ordered $d$-simplex $\Delta_d^*$, it is automatically determined on all of $\Delta_d$ by symmetry. It is not really difficult to prove this, but it is perhaps best seen by considering a few simple examples.

**Example 4.2.** Consider, first, the case of a coin which we know is biased 2:1 in favor of one side, but where we don't know which side it is - it could be either with equal probability. Then, $p_1^* = 2/3$ and $p_2^* = 1/3$. In terms of the original, unordered probabilities, this corresponds to either $p_1 = 2/3, p_2 = 1/3$ or $p_1 = 1/3, p_2 = 2/3$ and, since we are indifferent between categories, these two possibilites are equally likely; thus, we have as the mixing measure on the simplex $\Delta_2$ the measure

$$dF(p) = \frac{1}{2}\delta_{2/3} + \frac{1}{2}\delta_{1/3},$$

where $\delta_x$ is the Dirac measure which assigns probability 1 to $x$. This is a partition exchangeable probability , since it is invariant under the interchange of outcomes $H \to T$ & $T \to H$. △

**Example 4.3.** Consider next a die with six faces. The most general exchangeable probability is obtained by mixing multinomial probabilities over the simplex $\Delta_6$. The partition exchangeable probabilities are those which are invariant with respect to interchange of the faces. This would be equivalent to specifying a probability over

$$\Delta_6^* =: \{(p_1^*, \ldots, p_6^*) : p_1^* \geq \cdots \geq p_6^* \geq 0 \quad \& \quad \sum_{j=1}^{6} p_j^* = 1\}.$$

Specifying such a probability would be to say we have opinions about the bias of the die (the most likely face has probability $p_1^*$, the second most likely $p_2^*$, and so on), but not about *which* face has the bias, since our probability function is symmetric with respect to faces. A little thought should make it clear that the frequencies of frequencies can provide information relevant to the prior $dF$ on $\Delta_d^*$ (in the partition exchangeable case). For example, suppose that we know the die is biased in favor of one face, and that the other faces are equally likely. Then the unknown vector of ordered probabilities satisfies $p_1^* > p_2^* = \cdots = p_6^*$. Suppose now that in 100 trials we observe the frequency vector $(20, 16, 16, 16, 16, 16)$. Then we would guess approximately that $p_1 = p_1^* = .2$, and $p_2 = p_2^* = p_3 \cdots = p_6^* = .16$. But, if we observed the frequency vector (20, 40, 10, 15, 10, 5), we would guess $p_2 = p_1^* = .4$, and $p_1 = p_2^* = (20+10+15+10+5)/\{(100)(5)\} = .12$. Our estimate for $p_1$ differs in the two cases (.16 vs. .12) despite the fact the frequency count for the first category is the same in both cases. △

In general, the predictive probabilities for partition exchangeable probabilities will have the form[5]

$$p(X_{N+1} = c_i | X_1, \ldots, X_N) = f(n_i, a_0, a_1, \ldots, a_N).$$

## 4.4 Exchangeable Random Partitions

In this chapter we address the challenge we contemplated in the beginning. How can a Bayesian allow for *a*) infinite categories, or *b*) unknown species? If the number of categories is infinite, then no prior can be category symmetric, for such a prior would have to assign equal weight to each category, which is impossible; that is, if there are an infinite number of colors say, $c_1, c_2, \ldots$, then

$$p(X_1 = c_1) = p(X_2 = c_2) = \cdots = 1/d,$$

which is impossible, since $d = \infty$. We are thus compelled to consider probability assignments which contain some element of asymmetry between the different categories.

Let's us consider the problem more precisely; *what* does it mean to assign probabilities in a situation where we are encountering previously unknown species, continuously observing new and possibly not suspected kinds? According to the classical Bayesian reasoning, one assigns probabilities in advance to all possible outcomes and, then, updates via Bayes's theorem as new information comes in. How can one hypothesize and assign probabilities when the possible outcomes are unknown beforehand?

The earlier discussion of partition exchangeable sequences suggests a solution to this second difficulty. Rather than focus on the probability of a sequence of outcomes $(e_1, e_2, \ldots, e_N)$, or the probability of a frequency vector $(n_1, n_2, \ldots, n_N)$, focus instead on the partition vector $(a_1, a_2, \ldots, a_N)$ and its probability. Even if one does not know *which* species are present prior to sampling, one can still have beliefs as to the *relative abundances* in which those species, as yet unobserved, will occur[6]. One can now proceed exclusively at the level of partition vectors, and construct a theory of the type we are looking after.

What we should bear in mind cautiously in sampling of species scenario is that the relevant information being received is an *exchangeable random partition.*

[5]Johnson's sufficientness postulate thus makes the very strong supposition that the predictive probabilities reduce to a function $f(n_i, N)$. For also somehow weaker assumption than Johnson's, that result in a functionality of the predictive probability as $f(n_i, a_0, N)$ see [29].

[6]Note that in this setting $a_0$ is excluded from the partition vector and the reason why is, because of lacking prior knowledge as to the totality of species present, it is impossible to specify at any given stage how many species present since some of them that we are not aware of their size are yet to appear in the sample.

Because the individual species do not, in effect, have an individuality - we simply observe the first species, then at some subsequent time the second, at a still later time the third, and so on - the relevant information being received is a *partition* of the integers. In other words, the first species occurs at some set of times

$$A_1 =: \{d_1^1, d_1^2, d_1^3, \ldots : d_1^1 < d_1^2 < d_1^3 < \cdots\},$$

where necessarily $d_1^1 = 1$, and in general the set $A_1$ may only contain a finite number of times even if an infinite number of observations is made[7]. Likewise, the second species occurs at some set of times

$$A_2 =: \{d_2^1, d_2^2, d_2^3, \ldots : d_2^1 < d_2^2 < d_2^3 < \cdots\},$$

where necessarily $d_2^1$ is the first positive integer not in $A_1$, and $A_2$ may again be finite. In general, the $i$th species to be observed occurs at some set of times $A_i = \{\mathbf{d}_i^j : j = 1, 2, 3, \ldots\}$ and the collection of sets $A_1, A_2, A_3, \ldots$ forms a *partition* of the positive integers $\mathbf{N}$ in the sense that

$$\mathbf{N} = A_1 \cup A_2 \cup A_3 \cdots \quad \& \quad A_i \cap A_j = \emptyset, \quad \forall\, i \neq j.$$

In the example 4.1, the partition of $\{1, 2, 3, \ldots 10\}$ generated is

$$\{1, 7\} \cup \{2, 5, 10\} \cup \{3\} \cup \{4, 8, 9\} \cup \{6\}.$$

Note a new interpretation we can now give the partition vector $\mathbf{a} = (a_1, a_2, \ldots, a_{10})$; it records the sizes of the sets in the partition and the number of species observed. Thus, in Example 4.1, given the partition vector (2,1,2,0,...,0), two sets in the resulting partition have a single element [8], one set in the partition has two elements [9], two sets in the partition have three elements [10], and the total number of species observed is[11] 5. Although originally defined in terms of the underlying sequence, the partition vector is a function solely of the resulting partition of the time set; and one can therefore refer to the partition vector of a partition.

Thus, observing the successive species in our sample generates a random partition of the positive integers. Now let us consider in what sense such a partition could be 'exchangeable'. An obvious idea is to examine the structure of random partition arising from exchangeable sequences and see if we can characterize them in some way.

---

[7]This will happen if the first species is only observed a finite number of times, possibly even only once, in which case $A_1 = \{d_1^1\}$.

[8]Since $a_1 = 2$.

[9]Since $a_2 = 1$.

[10]Since $a_3 = 2$.

[11]Since $a_1 + a_2 + \cdots + a_{10} = 5$.

This turns out to be relatively simple: *if a random sequence is exchangeable, then the partition structures for two possible sequences have the same probability whenever they have the same partition vector* **a**.

In order to see this, let's think about what happens to a partition when we permute the categories or times of the underlying sequence which generates it. Consider our earlier example of the sequence {5,2,6,1,2,3,5,1,1,2}, and suppose we permute the category index in some way, say, the cyclic permutation

$$1 \to 2 \to 3 \to 4 \to 5 \to 6 \to 1.$$

Then, our original sequence becomes transformed into {6,3,1,2,3,4,6,2,2,3}, and the resulting partition of the time set from 1 to 10 is the same as before: species 6 occurs at times 1 and 7, hence, we get $A_1 = \{1, 7\}$, and so on. *Permuting the category index results in a new sequence but leaves the resulting partition unchanged.* Next, suppose we were to permute the times, say, by the cyclic permutation

$$1 \to 2 \to 3 \to 4 \to 5 \to 6 \to 7 \to 8 \to 9 \to 10 \to 1,$$

that is , what happened at time 1 is observed at time 2 instead; at time 2, at time 3 instead; and so on. Then, our original sequence becomes transformed into {2,5,2,6,1,2,3,5,1,1,}, and we get a new partition of the time set, namely,

$$\{1, 3, 6\} \cup \{2, 8\} \cup \{4\} \cup \{5, 9, 10\} \cup \{7\}.$$

Because of the exchangeability of the underlying sequence, this new partition has the same probability of occuring as the original one. Note that it has the same frequency vector **n** and, therefore, partition vector **a**. This observation is the one underlying the idea of an exchangeable random partition:

**Definition 5.** *A random partition is* ***exchangeable*** *if any two partitions $\pi_1$ and $\pi_2$ having the same partition vector have the same probability; i.e., if*

$$\boldsymbol{a}(\pi_1) = \boldsymbol{a}(\pi_2) \Rightarrow p(\pi_1) = p(\pi_2).$$

### 4.4.1 The Kingman representation theorem

In the case of sequence, the de Finetti representation theorem states that in the case of *i.i.d.* random variables, the general exchangeable sequence can be constructed out of elementary building blocks *i.e* Bernoulli trials in the case of 0,1-valued random variables and multinomial trials in the case of $d$-valued random variables. The corresponding building blocks of the general exchangeable random partition are the *paintbox processes.* In order to construct a paintbox process, consider an ordered 'defective' probability vector

$$\mathbf{p} = (p_1, p_2, \ldots), \text{ where } p_1 \geq p_2 \geq \ldots \geq 0 \text{ and } p_1 + p_2 + \cdots + \leq 1,$$

and let $\nabla$ denote the finite simplex of all such vectors. Now given such a defective probability vector $\mathbf{p} = (p_1, p_2, \ldots) \in \nabla$, let $p_0 = 1 - \sum_i p_i$; and let $F_{\mathbf{p}}$ be a probability measure on the unit interval [0,1] having point masses $p_j$ at some set of distinct points $x_j$, $j \geq 1$, and a continuous component assigning mass $p_0$ to [0,1]. Call such a probability measure a *representing probability measure* for $\mathbf{p}$ as in [53]. Let $X_1, X_2, \ldots$ be a sequence of *i.i.d* random variables with common distributions $F_{\mathbf{p}}$, and consider the exchangeable random partition generated by the rule

$$A_j = \{i : X_i = x_j\} \text{ where } A_1 \cup A_2 \cup \cdots = \{1, 2, \ldots, N\}.$$

That is, partition the integers 1,2, $\ldots$, $N$ by grouping together those times $i$ at which the random variables $X_j$ have a common value $x_j$. Then if $\mathbf{p} \in \nabla$, and $F_{\mathbf{p}}$ and $G_{\mathbf{p}}$ are two different representing probability measures for $\mathbf{p}$, then $F_{\mathbf{p}}$ and $G_{\mathbf{p}}$ generate the same exchangeable random partition $\Pi$, in the sense that the two random partitions have the same stochastic structure [53]. Thus, we have a well-defined rule for associating exchangeable random partitions with vectors in $\nabla$: given $\mathbf{p}$, select $F_{\mathbf{p}}$ and use $F_{\mathbf{p}}$ to generate $\Pi$. Let us call this resulting exchangeable random partition $\Pi_{\mathbf{p}}$. Now we are ready to for the Kingman representation theorem that we restate here without explicit proof. For details one can refer to [38].

**Theorem 6.** *The general exchangeable random partition is a mixture of paintbox processes.*

We can illuminate this further. Suppose that $Z_1, Z_2, Z_3, \ldots$ is a sequence of random partitions; specifically, for each $N \geq 1, Z_N$ is an exchangeable random partition of $\{1, 2, \ldots, N\}$. There is an obvious sense in which such a sequence is consistent. Namely, any partition of $\{1, 2, \ldots, N+1\}$ gives rise to a partition of $\{1, 2, \ldots, N\}$ by simply omitting the integer $N+1$ from the subset in which it occurs. Let $T_{N+1,N}$ denote the map which performs this operation. Then, the pair $Z_{N+1}$ and $Z_N$ are *consistent* if

$$p(Z_N \in A) = p(T_{N+1,N}(Z_{N+1}) \in A),$$

where $A$ is a set of partitions of $\{1, 2, \ldots N\}$; and the sequence is consistent if $Z_N$ and $Z_{N+1}$ are consistent for every $N \geq 1$. Every such consistent sequence gives rise to a probability measure on the partitions of $N = \{1, 2, \ldots\}$. If $\Pi$ is the probability distribution on the partitions of $N$ arising from such an arbitrary exchangeable random partition, then the Kingman representation theorem states that there exists a unique probability measure $dF$ on $\nabla$, the finite simplex of all ordered defective probability vectors, such that for every measurable set $A$ of partitions,

$$\Pi(A) = \int_{\nabla} \Pi_p(A) dF(\mathbf{p}).$$

Note that instead of integrating over the probability simplex, one integrates over the ordered defective probability simplex $\nabla$ consisting of all possible defective probability vectors $\mathbf{p}$. Moreover, as proven by Kingman[12], the ordered sample frequencies arising from the random partition converge in joint distribution to the mixing measure[13] $dF$.

### 4.4.2 The Poisson-Dirichlet process

The classical theories of induction that employ probability theory usually attempt to go further and identify classes of possible priors $dF$ thought to represent situations of limited information. In the de Finetti representation case discussed earlier, this was easy. The so-called flat priors $dp$ or $dp_1 dp_2 \ldots dp_{d-1}$ immediately suggested themselves, and the rest of the attempt was to come up with characterization of these priors in terms of symmetry assumptions about the underlying cylinder set probabilities. Here, however, it is far from apparent what a 'flat' prior would be.

Here we present the idea of Kingman. Let $\alpha > 0$. Suppose we took a symmetric Dirichlet prior $\mathbf{D}(\alpha)$ on the $d$-simplex $\Delta_d$ and let the number of categories tend to infinity, *i.e.*, $d \to \infty$. The resulting probabilities would then wash out. For any fixed $d_0 < \infty$ and $(x_1, x_2, \ldots, x_{d_0}) \in \Delta_d$, the cylinder set probabilities are

$$p_{\alpha,d}(p_1 \leq x_1, p_2 \leq x_2, \ldots, p_{d_0} \leq x_{d_0}) \to 0 \quad as \quad d \to \infty.$$

But, suppose instead that we consider the vector of ordered probabilities. Then, this provides us with the necessary tool. Since we can map the $d$-simplex $\Delta_d$ onto the ordered $d$-simplex[14] $\Delta_d^*$, the symmetric Dirichlet prior on $\Delta_d$ induces a probability distribution on $\Delta_d^*$. Since for any fixed $d_0 \leq d < \infty$ and sequence $(x_1^*, x_2^*, \ldots, x_{d_0}^* \in \Delta_d^*)$, there is a corresponding cylinder set probability

$$p_{\alpha,d}(p_1^* \leq x_1^*, p_2^* \leq x_2^*, \ldots, p_{d_0}^* \leq x_{d_0}^*).$$

[12]The distinctive role that the *continuous* component $p_0$ of a paintbox process plays in the theorem deserves some comment. When Kingman first investigated exchangeable random partitions, he was puzzled by the fact that mixtures over the *discrete* nondefective ordered probabilities $(p_1^*, p_2^*, \ldots)$ generated many, but by no means all possible exchangeable random partitions. The key to this puzzle is the far from obvious observation that when a new species appears, it is condemned to one of these scenarios: either it never appears again, or it is subsequently seen an infinite number of times. No intermediate case is possible. The species that arise once and only once are precisely those that arise from the continuous component. The Reverend Dr. Richard Price on the other hand, would not have found this surprising. As he stated, the first appearance of an event only informs us of its *possibility*, but would not "give us the least reason to apprehend that it was, in that instance or in any other, regular rather than irregular in its operations"; that is, we are given no reason to think that its probability of recurring is positive (read 'regular') rather than 0 (read 'irregular'). In effect, Price is saying that the first observation tells us that the outcome lies in the *support* of the unknown representing probability $F_p$, while the second observation tells us that it lies in the *discrete component* of this probability. For more discussion consult [3].

[13]Just as in de Finetti's theorem the *unordered* sample frequencies $(\frac{n_1}{N}, \ldots, \frac{n_d}{N})$ converge to the mixing measure $dF$, in the de Finetti representation, here the *ordered* sample frequencies converge to the mixing measure $dF$ in the Kingman representation.

Then, as Kingman shows, if $d \to \infty$ *and* $\alpha \to 0$ in such a way that $d\alpha \to \theta > 0$, for some positive number $\theta$, then the resulting sequence of probabilities does not 'wash out': instead, it has a proper limiting distribution. And, since this is so for each $d$, the result is a probability measure[15] on $\nabla$. This is called the *Poisson-Dirichlet distribution*[16] (with parameter $\theta$). A simple example will illustrate the phenomenon. Suppose you pick a point $\mathbf{p}$ at random from $\Delta_d$ according to the symmetric Dirichlet distribution $p_{\alpha,d}$ and ask for the probability $p_{\alpha,d}(p_1 \geq x_1)$. As $d \to \infty$, this probability tends to 0 (since a typical coordinate of $\mathbf{p}$ will be small if $d$ is large). But suppose, instead, you ask for the probability that the *maximum* coordinate of $\mathbf{p}$ exceeds $x_1$: that is, $p_{\alpha,d}(p_1^* \geq x_1)$. Then, Kingman's theorem states that this probability has a *nonzero* limit as $d \to \infty$.

### 4.4.3 The Ewens sampling formula

Since the Poisson-Dirichlet distribution with parameter $\theta$ is a probability measure on $\nabla$, and each paintbox process in $\nabla$ gives rise to an exchangeable random partition, for every sample size $N$ the Poisson-Dirichlet distribution induces a probability distribution $p(a_1, a_2, \ldots, a_N)$ on the set of possible partition vectors. Kingman shows that these probabilities are given by the so-called Ewens sampling formula[17],

$$\textbf{Ewens Sampling Formula:} \quad \frac{n!}{\theta(\theta+1)\ldots(\theta+n-1)} \prod_{r=1}^{n} \frac{\theta^{a_r}}{r^{a_r} a_r!}.$$

Given the Ewens formula for the cylinder set probabilities, it is a simple task to derive the corresponding predictive probabilities or rules of succession. It is important, however, to be clear what this means, so an additional clarification is useful. Suppose we are performing a sequence of observations $X_1, X_2, \ldots, X_N, \ldots$, noting in each stage the species of an animal for example. At each point then, we observe either a species previously observed or an entirely new species. Before these are observed, it doesn't make sense to refer to these outcomes as exchangeable; in fact, it doesn't even make sense to refer to the probabilities of such outcomes. Because ahead of time we don't know what a complete list of possible outcomes is. We are hence learning as we go along. But at time $N$ we can construct a partition of $\{1, 2, \ldots, N\}$ on the basis of what we have seen thus far, and it *does* make sense to talk prospectively about the probability of seeing a particular partition. It is then natural to assume that the resulting random partition is exchangeable. Having arrived at this state, we can then invoke the Kingman representation theorem, and write our exchangeable random partition as a mixture of paintbox processes. Although we do not have prior beliefs about the probabilities of the species we observe, we can certainly have opinions about their *abundances*, *i.e.*, what is the frequency of occurrence of the most abundant species, the second most abundant, and so on, and this is what our prior on $\nabla$ summarizes.

[14]By association to any vector $(p_1, p_2, \ldots, p_d)$ its ordered rearrangement $(p_1^*, p_2^*, \ldots, p_d^*)$.

[15]A 'consistent' set of probabilities on the finite cylinder sets always corresponds to a unique probability on infinite sequence space.

[16]The terminology is intended to suggest an analogy with the classical Poisson-binomial limit theorem in probability theory.

[17]This formula proved to be highly ubiquitous; it crops up in a large number of seemingly unrelated contexts. One example of many is: if one picks a random permutation of the integers {1,2, …, N}, and lets $a_j$ denote the number of $j$-cycles, then the probability distribution for $a_1, a_2, \ldots, a_N$ is provided by the Ewens formula.

Now, given that we make a series of $N$ observations, it is clear that our exchangeable probability assignment will predict whether a new species will be observed on the next trial. And, if we don't observe a new species, whether we see a member of the same species as the very first animal observed[18]. Or, whether a member of the second species observed[19].

Lastly we will give credit to the De Morgan's intuition as we discussed earlier. Suppose that we have observed a number of species so far with $n_1$ of the first type, $n_2$ of the second, and so on. Then the succession probabilities for observing one of the known species or an unknown one, given the Poisson-Dirichlet prior (and lettings$s_j$ denote the $j$th species observed to date), is [53]

$$p(X_{N+1} = s_j|\mathbf{n}) = \frac{n_j}{(N+\theta)}$$

that is, with $\theta = 1$ and $d = 0$ the answer is identical to De Morgan's. This implicates that De Morgan's derivation was not an arbitrary hypothesis, but a mathematical formula with a concrete background.

[18]This is the same as whether the new partition resulting after time $N+1$ will add the integer $N+1$ to the member of the partition containing 1.

[19]This is the same as whether the new partition adds $N+1$ to that member of the partition containing the first integer not in the member of the partition containing 1.

# Chapter 5

# Predictive Classification II

So far we have equipped ourselves with the necessary tools to tackle the classification task based on some discrete features and also in the latter chapter we investigated the type of exchangeability that we need to handle the case of classification with some unknown outcomes of the features variable. Here in this chapter, we demonstrate that classifiers based on partition exchangeability behave differently from classifiers derived under de Finetti type exchangeability discussed in chapter three, where it was concluded that marginal and simultaneous predictive classifiers of test data become congruent almost surely when the amount of training data tends to infinity. Under partition exchangeability however, a positive probability always remains for the event that marginal and simultaneous predictive classifiers do not coincide, such that the population of test data items will play a clear role when assessing how surprising any particular observed event is.

This chapter is structured as follow. In first section we consider the behaviour of supervised predictive classifiers under partition exchangeability in relation to generic convergence results in predictive inference, while classifiers under partition exchangeability are introduced and their properties are examined in second section.

## 5.1 Supervised Predictive Classification

In harmony with the previous notation, we denote the set of $m$ available training items by $M$ and correspondingly the set of $n$ test items by $N$. For each item we observe a finite vector of $d$ features, such that the element for the feature $j$ takes values in an alphabet $\mathcal{X}_j, j = 1, ..., d$. In particular, these alphabets are not assumed known or fixed *a priori*, or even after observing the training items, such that the test items may contain feature values not previously seen. Hence, this special property poses a challenge for the inductive inference, as an assumption of a fixed alphabet for a feature would necessitate a retrospective change of the predictive probabilities for all previous items whenever a new test item with a previously unobserved feature value appears.

As discussed in chapter four, a solution to the problem with induction in this setting arises by an application of the theory of partition exchangeability [40], which acknowledges the presence of previously unanticipated phenomena when predictive probabilities are formed from data[1]. Partition exchangeability has been used in the context of genetic data [16],[15], however, [11] defines generative supervised classification models based on this theory and this chapter is mainly based on this reference. For notational simplicity, we will consider each feature alphabet as the set $\mathcal{X}_j = \{1, \ldots, r_j\}$, where $r_j$ is a positive, non-fixed integer remaining implicit in our analysis for all $j = 1, \ldots, d$. A training item $i \in M$ is then characterized by a feature vector $\mathbf{z}_i$, with the elements $z_{ij} \in \mathcal{X}_j, j = 1, ..., d$. Similarly, we have for a test item $i \in N$ the feature vector $\mathbf{x}_i$, with elements $x_{ij} \in \mathcal{X}_j$. Collections of the training and test data vectors are denoted by $\mathbf{z}^{(M)}$ and $\mathbf{x}^{(N)}$, respectively.

Let the training data represent $k$ classes, such that $T$ is a joint labeling of all the training items. In simultaneous supervised classification we will assign labels to all the items in $N$ in a joint fashion. A joint labeling $S$ remains even here as an ordered partition of $N$ such that the classes are identified by those present in $T$. Since any particular subset of classes present in $T$ must be allowed to lack test items, some classes can remain empty in $S$, in which case the actual number of classes in $S$ is $1 \leq k' \leq k$. A structure $S$ implies a partition of the feature vectors, such that $\mathbf{x}^{(s_c)}, s_c \subseteq N$, represents the collection of data for the items in class $c = 1, ..., k$. Let $\mathcal{S}$ again denote the space of eligible simultaneous classification structures for a given $N$.

Using a stochastic urn model for supervised classification [13], a prior probability $p(S|T)$ of a classification structure $S$ is obtained conditionally on the known fixed classification $T$ of the training data. Here we simply assume that the prior $p(S|T)$ is the uniform distribution over $\mathcal{S}$ as discussed in chapter three. Given a prior and the collections of training and test data, a simultaneous classifier is based on the posterior distribution over $\mathcal{S}$, which is defined as (3.4)

$$p(S|\mathbf{x}^{(N)}, \mathbf{z}^{(M)}, T) = \frac{p(\mathbf{x}^{(N)}|\mathbf{z}^{(M)}, S, T)p(S|T)}{\sum_{S \in \mathcal{S}} p(\mathbf{x}^{(N)}|\mathbf{z}^{(M)}, S, T)p(S|T)}, \tag{5.1}$$

where $p(\mathbf{x}^{(N)}|\mathbf{z}^{(M)}, S, T)$ is the conditional predictive probability for the entire observed population of test data. In contrast, a marginal classifier specifies the predictive probabilities independently for each test item, as already presented in (3.31). Here we will explore how the simultaneous and marginal forms of inductive inference will relate to each other under the circumstances where alphabets of discrete-valued features are not fixed *a priori.*

[1]This type of exchangeability is what we only consider in this chapter and all its sections. Reader can consider the suffix 'under partition exchangeability ' latently in the title of this chapter's sections.

For the purpose of later examination of predictive probabilities, we will consider separately the process of establishing sufficient statistics from the data $\mathbf{x}^{(N)}$ alone and jointly from $\mathbf{z}^{(M)}$ and $\mathbf{x}^{(N)}$, conditional on $S$ and $T$. Given a structure $S$, sufficient statistics arise under *partition exchangeability* for each subset of data $\mathbf{x}^{(s_c)}$. More specifically, we assume an unrestricted form of exchangeability [13],[15] which implies a product predictive measure over the $d$ features. The operative interpretation of this form of exchangeability is that the features are modelled as conditionally independent given $S$.

Given a structure $S$ with $k$ classes, a predictive model is obtained by assuming exchangeable equivalence relations for each feature in each class [15],[37]. For counting purposes, we assign indices to items in $s_c$ using an arbitrary permutation of integers $1, ..., |s_c|$. To simplify the notation we use $n_c = |s_c|$ to denote the cardinality of a class in the sequel. Let $I(\cdot)$ be an indicator function and $l$ index a value in the alphabet $\mathcal{X}_j$. Define $n_{cjl} = \sum_{i \in s_c} I(x_{ij} = l)$ as the frequency of items in class $c$ that carried the value $l$ for the feature $j$. Then, the information in $\mathbf{x}^{(s_c)}$ can be compressed for each feature in terms of the count

$$\rho_{cjt} = \sum_{l=1}^{\infty} I(n_{cjl} = t), \tag{5.2}$$

which defines the frequency of distinct feature values that have been observed exactly $t$ times for feature $j$ in class $c$. Further, let $\rho_{cj} = (\rho_{cjt})_{t=1}^{n_c}$ denote a partition of the integer $n_c$ into a vector of such counts. In a series of works by John Kingman, e.g. [39], an explicit predictive model for observed values of single feature of the above type was established using the Definition 5. Here we restate the definition with defined notation. "*A random partition $\rho_{cj}$ is said to be exchangeable, if any two partitions of $n_c$ having the same partition vector, have the same predictive probability $p(\rho_{cj})$.*"

Kingman's representation theorem establishes that the predictive distribution of the observations for a single feature equals under the partition exchangeability a Poisson-Dirichlet($\psi$) distribution with parameter[2] $\psi$

$$p(\rho_{cj}) = \frac{n_c!}{\psi(\psi+1)\cdots(\psi+n_c-1)} \prod_{t=1}^{n_c} \left\{ (\frac{\psi}{t})^{\rho_{cjt}} \frac{1}{\rho_{cjt}!} \right\}. \tag{5.3}$$

As noted earlier, this particular probability result is also known as the Ewens sampling formula and it arises as an infinite mixture of probability measures over an infinite-dimensional simplex. Corander et al [11], by assuming that the partitions for distinct classes and features are unrestrictedly exchangeable given $S$, obtained the following product predictive measure for the entire collection of test data

$$p(\mathbf{x}^{(N)}|S) = \prod_{c=1}^{k}\prod_{j=1}^{d} \frac{n_c!}{\psi(\psi+1)\cdots(\psi+n_c-1)} \prod_{t=1}^{n_c} \left\{ (\frac{\psi}{t})^{\rho_{cjt}} \frac{1}{\rho_{cjt}!} \right\}. \tag{5.4}$$

[2]Note here again the change in notation, as $\psi$ was earlier used to parametrise the label generating process.

We now turn our attention to the inductive inference process where predictions about $\mathbf{x}^{(N)}$ are also conditioned on the training data $\mathbf{z}^{(M)}$ and their *a priori* given labelling $T$. Given that $m_c$ and $m_{cjl}$ are defined analogously to $n_c$ and $n_{cjl}$, respectively, we obtain the partition vector $\tilde{\boldsymbol{\rho}}_{cj} = (\rho_{cjt})_{t=1}^{n_c+m_c}$ with elements given by

$$\tilde{\rho}_{cjt} = \sum_{l=1}^{\infty} I(n_{cjl} + m_{cjl} = t). \tag{5.5}$$

Using Bayes' theorem and the above result based on the Kingman's representation theorem, the predictive probability of the test data supervised by the training data and the two joint labellings $S, T$ can be written as [11]

$$p(\mathbf{x}^{(N)}|\mathbf{z}^{(M)}, S, T) = \frac{p(\mathbf{x}^{(N)}, \mathbf{z}^{(M)}|S, T)}{p(\mathbf{z}^{(M)}|T)} = \tag{5.6}$$

$$\frac{\prod_{c=1}^{k}\prod_{j=1}^{d} \frac{(m_c+n_c)!}{\psi(\psi+1)\cdots(\psi+m_c+n_c-1)} \prod_{t=1}^{m_c+n_c} \left\{ (\frac{\psi}{t})^{\tilde{\rho}_{cjt}} \frac{1}{\tilde{\rho}_{cjt}!} \right\}}{\prod_{c=1}^{k}\prod_{j=1}^{d} \frac{m_c!}{\psi(\psi+1)\cdots(\psi+m_c-1)} \prod_{t=1}^{m_c} \left\{ (\frac{\psi}{t})^{\rho_{cjt}} \frac{1}{\rho_{cjt}!} \right\}}$$

$$= \prod_{c=1}^{k}\prod_{j=1}^{d} \frac{\frac{(m_c+n_c)!}{m_c!}}{(\psi+m_c)(\psi+m_c+1)\cdots(\psi+m_c+n_c-1)} \frac{\prod_{t=1}^{m_c+n_c} \left\{ (\frac{\psi}{t})^{\tilde{\rho}_{cjt}} \frac{1}{\tilde{\rho}_{cjt}!} \right\}}{\prod_{t=1}^{m_c} \left\{ (\frac{\psi}{t})^{\rho_{cjt}} \frac{1}{\rho_{cjt}!} \right\}}.$$

## 5.2 Asymptotic Properties of Supervised Classifiers

We will now examine how the predictive probabilities for simultaneous and marginal supervised classifiers introduced in the previous section relate to each other in the presence of increasing amounts of training data. As discussed earlier, these two predictive probabilities need not in general be equal, even if the test data were generated *i.i.d.* from the same underlying distribution as the training data. The asymptotic equality of simultaneous and marginal classifiers with respect to an increasing amount of available training data was established in chapter three under de Finetti type of exchangeability. Here we show that this property does not hold under the classification model arising from the partition exchangeability. An intuitive implication of this result is that the simultaneous classifier enjoys the advantage of jointly modelling the entire population of labels of test items, such that it can better assess the level of surprise in any particular observed event in relation to other events. Moreover, this effect does not wear out almost surely even if the amount of training data tends to infinity.

We need to introduce some additional notation for the marginal classifier. Let ${\tilde{\rho}_{cjt}}^{(i)}$ denote the updated sufficient statistic (see eq. (5.5)) when data $\mathbf{x}_i$ from only a single test item is taken into account, defined as ${\tilde{\rho}_{cjt}}^{(i)} = \sum_{l=1}^{\infty} I(m_{cjl} + n_{i;cjl} = t), t = 1, \ldots, m_c + 1$, where $n_{i;cjl}$ is the observed frequency of category $l$ for feature $j$ in item $i$. Thus, there is only a single value $l_j^{(i)}$, such that $n_{i;cjl_j^{(i)}} = 1$, and $n_{i;cjl_j^{(i)}} = 0$, otherwise. Consequently, we can rewrite ${\tilde{\rho}_{cjt}}^{(i)}$ as

$$(5.7) \quad {\tilde{\rho}_{cjt}}^{(i)} = \sum_{l=1}^{\infty} I(m_{cjl} + n_{i;cjl} = t) = \rho_{cjt} - I(m_{cjl_j^{(i)}} = t) + I(m_{cjl_j^{(i)}} = t - 1).$$

We would further see, if $m_{cjl^{(i)}} > 0$, there is always a single frequency $t_j^{(i)}$, such that $m_{cjl^{(i)}} = t_j^{(i)}$ and ${\tilde{\rho}_{cjt}}^{(i)} = \rho_{cjt}$ for all $t \neq t_j^{(i)}$ and $t \neq t_j^{(i)} + 1$. In cases where $m_{cjl^{(i)}} = 0$, item $i$ carries a unique value for feature $j$ and the updating of sufficient statistics behaves slightly differently in the following two cases. If no other unique values are present in class $c$ for this feature, then ${\tilde{\rho}_{cjt}}^{(i)} = \rho_{cjt}$ for all $t > 1$ and ${\tilde{\rho}_{cj1}}^{(i)} = 1$. If there are multiple unique values present in class $c$ for this feature, then ${\tilde{\rho}_{cjt}}^{(i)} = \rho_{cjt}$ for all $t > 1$ and ${\tilde{\rho}_{cj1}}^{(i)} = \rho_{cj1}+1$. In order to avoid burdening the notation excessively, we will not explicitly consider this distinction in the lemma below. Note that the above definition of ${\tilde{\rho}_{cjt}}^{(i)}$ still holds in all these different cases.

To compare the predictive probabilities of simultaneous and marginal classifiers, we now consider the case with $S = \mathring{S}$, where the latter denotes the collection of the labels for all single test items that are each considered only in terms of $\mathbf{x}_i, \mathbf{z}^{(M)}, T$, and $S_i$. The following lemma shows how the two probabilities are related to each other when the amount of training data increases.

**Lemma 1**. *Asymptotic behavior of the difference of log predictive probabilities for simultaneous and marginal classifiers under partition exchangeability. Let* $m_c$ *be* *very large for all $c = 1, ..., k$. Then*

$$\log p(\mathbf{x}^{(N)}|\mathbf{z}^{(M)}, S, T) - \log p(\mathbf{x}^{(N)}|\mathbf{z}^{(M)}, \mathring{S}, T)$$
$$\approx \sum_{c=1}^{k}\sum_{j=1}^{d}\left\{\sum_{t=1}^{m_c+n_c}\tilde{\rho}_{cjt}\log(\frac{\psi}{t}) - \sum_{t=1}^{m_c}\rho_{cjt}\log(\frac{\psi}{t}) + \sum_{i\in s_c}\log(\frac{t_j^{(i)}+1}{t_j^{(i)}})\right.$$
$$\left. - \sum_{t=1}^{m_c+n_c}\log(\tilde{\rho}_{cjt}!) + \sum_{t=1}^{m_c}\log(\rho_{cjt}!) - \sum_{i\in s_c}\log(\rho_{cjt_j^{(i)}}) + \sum_{i\in s_c}\log(\rho_{cj(t_j^{(i)}+1)})\right\}.$$

*Proof.* The proof is given in the Appendix A. □

The result in the above lemma highlights the difference between the simultaneous and marginal classifiers in terms of inductive reasoning. In the simultaneous classifier the log predictive probabilities contrast the changes of sufficient statistics between the training data and the entire population of test items as a whole. Instead, the marginal classifier sums the changes induced by each test item separately and therefore, it by definition never pays attention to how surprising any observed event is in relation to the events observed for the remaining population of test items. The following theorem excerpted from [11], shows explicitly that this difference can persist even in the presence of increasing amounts of training data.

**Theorem 7.** *Convergence of log predictive probabilities for simultaneous and marginal classifiers under partition exchangeability. For any $m \in \mathbb{Z}^+$, and any $m_c \geq m, c = 1, ..., k$, there exists $\epsilon > 0$ such that*

$$P[|\log p(\mathbf{x}^{(N)}|\mathbf{z}^{(M)}, S, T) - \log p(\mathbf{x}^{(N)}|\mathbf{z}^{(M)}, \mathring{S}, T)| > \epsilon] > 0.$$

*Proof.* The proof is given in the Appendix A. □

The following example illuminates this theorem numerically.

**Example 5.1.**

Consider the same training data as in example 3.1. In addition to our test data introduced there ($n = 3$), here we further consider 7 more test items such that within different features, we have observed new value categories marked by new integers. Further suppose that we have classified these items by an arbitrary classification rule so we are provided by their specific labels that denotes which class they belong to.

$$
(5.8)\qquad
\begin{array}{cccccc}
 & z_{i1} & z_{i2} & z_{i3} & z_{i4} & T \\
\mathbf{z}_1 & 1 & 2 & 1 & 3 & 1 \\
\mathbf{z}_2 & 1 & 2 & 2 & 3 & 1 \\
\mathbf{z}_3 & 2 & 1 & 2 & 1 & 1 \\
\mathbf{z}_4 & 3 & 3 & 3 & 1 & 1 \\
\mathbf{z}_5 & 3 & 3 & 1 & 2 & 1 \\
\mathbf{z}_6 & 1 & 3 & 2 & 2 & 2 \\
\mathbf{z}_7 & 2 & 3 & 1 & 1 & 2 \\
\mathbf{z}_8 & 2 & 2 & 3 & 3 & 2 \\
\mathbf{z}_9 & 1 & 2 & 2 & 3 & 2 \\
\mathbf{z}_{10} & 3 & 1 & 2 & 2 & 2
\end{array}
\qquad
\begin{array}{cccccc}
 & x_{i1} & x_{i2} & x_{i3} & x_{i4} & S \\
\mathbf{x}_1 & 1 & 2 & 3 & 1 & 1 \\
\mathbf{x}_2 & 1 & 1 & 2 & 1 & 1 \\
\mathbf{x}_3 & 2 & 3 & 1 & 1 & 1 \\
\mathbf{x}_4 & 1 & 3 & 4 & 3 & 1 \\
\mathbf{x}_5 & 1 & 4 & 5 & 1 & 1 \\
\mathbf{x}_6 & 1 & 3 & 3 & 3 & 2 \\
\mathbf{x}_7 & 4 & 2 & 3 & 1 & 2 \\
\mathbf{x}_8 & 5 & 1 & 3 & 4 & 2 \\
\mathbf{x}_9 & 6 & 4 & 2 & 3 & 2 \\
\mathbf{x}_{10} & 7 & 1 & 3 & 4 & 2
\end{array}
$$

Based on these data we want to calculate the predictive probability of the test data given in (5.6) and compare it with

$$
(5.9)\qquad p(\mathbf{x}^{(N)}|\mathbf{z}^{(M)}, \mathring{S}, T) = \prod_{c=1}^{k} \prod_{i:S_i=c} p(\mathbf{x}_i|\mathbf{z}^{(c)}, S_i = c, T),
$$

that $\mathring{S}$ is the overall labelling implied by the individual assignment of the labels to the items. For this, we need to obtain the sufficient statistics vector as given in (5.2) and (5.5).

$$
(5.10)\qquad
\boldsymbol{n} = \left\{
\begin{array}{rl}
\boldsymbol{n}_{11} = & (2,1,2,0,0) \\
\boldsymbol{n}_{12} = & (1,2,2,0,0) \\
\boldsymbol{n}_{13} = & (2,2,1,0,0) \\
\boldsymbol{n}_{14} = & (2,1,2,0,0) \\
\boldsymbol{n}_{21} = & (2,2,1,0,0) \\
\boldsymbol{n}_{22} = & (1,2,2,0,0) \\
\boldsymbol{n}_{23} = & (1,3,1,0,0) \\
\boldsymbol{n}_{24} = & (1,2,2,0,0)
\end{array}
\right\}
\qquad
\boldsymbol{\rho} = \left\{
\begin{array}{rl}
\boldsymbol{\rho}_{11} = & (1,2,0,0,0) \\
\boldsymbol{\rho}_{12} = & (1,2,0,0,0) \\
\boldsymbol{\rho}_{13} = & (1,2,0,0,0) \\
\boldsymbol{\rho}_{14} = & (1,2,0,0,0) \\
\boldsymbol{\rho}_{21} = & (1,2,0,0,0) \\
\boldsymbol{\rho}_{22} = & (1,2,0,0,0) \\
\boldsymbol{\rho}_{23} = & (2,0,1,0,0) \\
\boldsymbol{\rho}_{24} = & (1,2,0,0,0)
\end{array}
\right\}
$$

(5.11)

$$
\tilde{\boldsymbol{n}} = \left\{
\begin{array}{rl}
\tilde{\boldsymbol{n}}_{11} = & (6,2,2,0,0,0,0,0,0,0) \\
\tilde{\boldsymbol{n}}_{12} = & (2,3,4,1,0,0,0,0,0,0) \\
\tilde{\boldsymbol{n}}_{13} = & (3,3,2,1,1,0,0,0,0,0) \\
\tilde{\boldsymbol{n}}_{14} = & (5,2,3,0,0,0,0,0,0,0) \\
\tilde{\boldsymbol{n}}_{21} = & (3,2,1,1,1,1,1,0,0,0) \\
\tilde{\boldsymbol{n}}_{22} = & (2,3,4,1,0,0,0,0,0,0) \\
\tilde{\boldsymbol{n}}_{23} = & (2,4,4,0,0,0,0,0,0,0) \\
\tilde{\boldsymbol{n}}_{24} = & (2,2,4,2,0,0,0,0,0,0)
\end{array}
\right\}
\quad
\tilde{\boldsymbol{\rho}} = \left\{
\begin{array}{rl}
\tilde{\boldsymbol{\rho}}_{11} = & (0,2,0,0,0,1,0,0,0,0) \\
\tilde{\boldsymbol{\rho}}_{12} = & (1,1,1,1,0,0,0,0,0,0) \\
\tilde{\boldsymbol{\rho}}_{13} = & (2,1,2,0,0,0,0,0,0,0) \\
\tilde{\boldsymbol{\rho}}_{14} = & (0,1,1,0,1,0,0,0,0,0) \\
\tilde{\boldsymbol{\rho}}_{21} = & (5,1,1,0,0,0,0,0,0,0) \\
\tilde{\boldsymbol{\rho}}_{22} = & (1,1,1,1,0,0,0,0,0,0) \\
\tilde{\boldsymbol{\rho}}_{23} = & (0,1,0,2,0,0,0,0,0,0) \\
\tilde{\boldsymbol{\rho}}_{24} = & (0,3,1,0,0,0,0,0,0,0)
\end{array}
\right\}
$$

$$(5.12)\qquad \begin{array}{c} m = 10, \quad n = 10, \quad k = 2, \quad d = 4, \quad \psi = 5, \\ n_c = 5, \quad m_c = 5, \quad c = 1, 2, \quad j = 1, 2, \dots 4. \end{array}$$

With these information now we can invoke the (5.6) formula and obtain

$$p(\mathbf{x}^{(N)}|\mathbf{z}^{(M)}, S, T) = 1.757894 \times 10^{-8}$$

For calculating this value based on marginal classifier we need to obtain different $\tilde{\boldsymbol{\rho}}$ for each item independently. For example for $\mathbf{x}_7$ we have

(5.13)

$$\tilde{\boldsymbol{n}}^{(7)} = \left\{ \begin{array}{rl} \tilde{\boldsymbol{n}}_{11} = & (2,1,2,0,0) \\ \tilde{\boldsymbol{n}}_{12} = & (1,2,2,0,0) \\ \tilde{\boldsymbol{n}}_{13} = & (2,2,1,0,0) \\ \tilde{\boldsymbol{n}}_{14} = & (2,1,2,0,0) \\ \tilde{\boldsymbol{n}}_{21} = & (2,2,1,1,0,0) \\ \tilde{\boldsymbol{n}}_{22} = & (1,3,2,0,0,0) \\ \tilde{\boldsymbol{n}}_{23} = & (1,3,2,0,0,0) \\ \tilde{\boldsymbol{n}}_{24} = & (2,2,2,0,0,0) \end{array} \right\} \qquad \tilde{\boldsymbol{\rho}}^{(7)} = \left\{ \begin{array}{rl} \tilde{\boldsymbol{\rho}}_{11} = & (1,2,0,0,0) \\ \tilde{\boldsymbol{\rho}}_{12} = & (1,2,0,0,0) \\ \tilde{\boldsymbol{\rho}}_{13} = & (1,2,0,0,0) \\ \tilde{\boldsymbol{\rho}}_{14} = & (1,2,0,0,0) \\ \tilde{\boldsymbol{\rho}}_{21} = & (2,2,0,0,0,0) \\ \tilde{\boldsymbol{\rho}}_{22} = & (1,1,1,0,0,0) \\ \tilde{\boldsymbol{\rho}}_{23} = & (1,1,1,0,0,0) \\ \tilde{\boldsymbol{\rho}}_{24} = & (0,3,0,0,0,0) \end{array} \right\}$$

Then based on these vectors and (5.10), we can calculate $p(\mathbf{x}_i|\mathbf{z}^{(c)}, S_i = c, T)$ with (5.6) for $n = 1$, $i = 1, 2, \dots 10$, and multiply the values to obtain what is represented in (5.9).

$$p(\mathbf{x}^{(N)}|\mathbf{z}^{(M)}, \mathring{S}, T) = 2.636991 \times 10^{-6}$$

△

This example demonstrates explicitly how the predictive probabilities arise under the two alternative classification approaches.

# Chapter 6

# Conclusion and Remarks

In inductive classification, there are variety of ways one can adhere in classifying items to appropriate classes. While most of these strategies are based on Bayesian framework, its not an easy task to formally classify test items that are harbouring some new values or species. Nevertheless, retrospective changing of an alphabet upon arrival of new observations can be done, but this approach is not based on an inductive Bayesian framework that acknowledges the gradually updating predictions about the future in an autonomous manner. A truly inductive approach for prediction of a single future feature value using a hierarchical Dirichlet prior is introduced in [23], however, in a context different from classification.

We have discussed that an inductive supervised classification principle for predictable events corresponding to known alphabets or species, arises under the assumption of de Finetti type of exchangeability and that marginal and simultaneous classifiers based on this assumption will exhibit the same behaviour as the amount of training data tends to infinity. Moreover, in the case where the amount of test data increases instead, it was shown that the simultaneous classifier enjoys a limit where it saturates, such that no additional test data will improve the overall predictive ability. This property will not hold for the case of the presence of unanticipated events or unknown alphabets when the the size of the training data increases. The latter case, albeit arises under the assumption of partition exchangeability, since de Finetti type of exchangeability is unable to provide a framework that can be suitable for the cases where unpredictable feature values arise.

Moreover, for the inductive classification based on partition exchangeability, joint modelling of labels of a population of test items appears as a natural approach, since its capable of assessing the degree of surprise of an observed event for a test item in relation to both the test items and the training data. The basic rationale behind the use of joint classification models under de Finetti type exchangeability stems from the fact that labels perceived as random quantities, remain dependent even under an *i.i.d.* sampling model, as long as the generating probability measures for the classes are not exactly known. In contrast to a marginal classifier which predicts the label for each items individually based on an *i.i.d.* assumption, *i.e.* conditionally independent feature data given a fixed generating measure learned from the training items, a simultaneous classifier exploits the dependence which may have a considerable effect in the presence of sparse training data. A marginalized predictive classifier takes an additional step further for any single item by treating the labels of remaining items as nuisance parameters and deriving the predictive distribution of a label through its marginal posterior from the joint distribution of all the labels.

As already discussed, gains in classification accuracy are accessible through joint modelling of multiple test items and the optimality of such an approach has been considered by multiple authors[1] in different classification contexts. For example while [13] discusses the semi supervised classifiers for the case of known alphabets under de Finetti type exchangeability, exploiting the partition exchangeability for the cases with unpredictable alphabets in the semi supervised classification context is also interesting. It is plausible to consider and form new classes when the frequency of observing new species or alphabet values increases considerably. Another possible development would be to examine how the models in [23],[52] would behave asymptotically for a simultaneous prediction of a population of test items when supervised by an infinite amount of training data. It is currently an open question whether their asymptotic predictions would show similar characteristics as the predictions considered in last chapter under partition exchangeability, or if they would be reducible to a set of marginal predictions as under de Finetti type exchangeability [11]. Also it will be interesting to consider a log loss scoring as suggested in [20],[21].

[1] [11], [13],[14], [44], and [46].

# Appendices

# Appendix A

# Outlines of the Proofs of Theorems

These proofs are restatements of the results introduced in [11-13].

## A.1 Theorem 1

*Proof.* We recall first the logarithm of the predictive probability (3.20) for SPC which equals under the assumption stated in the theorem:
(A.1)

$$\log p(\mathbf{x}^{(N)}|\mathbf{z}^{(M)}, S, T) = \sum_{c=1}^{k}\sum_{j=1}^{d} \log \frac{\Gamma(m_c+1)}{\Gamma(n_c+m_c+1)} + \sum_{c=1}^{k}\sum_{j=1}^{d}\sum_{l=1}^{r_j} \log \frac{\Gamma(n_{cjl}+m_{cjl}+\lambda_{cjl})}{\Gamma(m_{cjl}+\lambda_{cjl})}.$$

Using Stirling's approximation on the gamma function terms in (A.1) it follows that,

$$\begin{aligned}
&\log p(\mathbf{x}^{(N)}|\mathbf{z}^{(M)}, S, T) = \\
&-\sum_{c=1}^{k}\sum_{j=1}^{d} \left(m_c + \frac{1}{2}\right) \log\left(1 + \frac{n_c}{m_c+1}\right) \\
&+\sum_{c=1}^{k}\sum_{j=1}^{d}\sum_{l=1}^{r_j} \left(m_{cjl} + \lambda_{cjl} - \frac{1}{2}\right) \log\left(1 + \frac{n_{cjl}}{m_{cjl}+\lambda_{cjl}}\right) \\
&-\sum_{c=1}^{k}\sum_{j=1}^{d} n_c \log(n_c + m_c + 1) \\
&+\sum_{c=1}^{k}\sum_{j=1}^{d}\sum_{l=1}^{r_j} n_{cjl} \log(n_{cjl} + m_{cjl} + \lambda_{cjl})
\end{aligned}$$

$$
\begin{aligned}
&= -\sum_{c=1}^{k}\sum_{j=1}^{d}\Big(m_c+\frac{1}{2}\Big)\Big(\frac{n_c}{m_c+1}-\frac{1}{2}\Big(\frac{n_c}{m_c+1}\Big)^2+\cdots\Big)\\
&\quad+\sum_{c=1}^{k}\sum_{j=1}^{d}\sum_{l=1}^{r_j}\Big(m_{cjl}+\lambda_{cjl}-\frac{1}{2}\Big)\Big(\frac{n_{cjl}}{m_{cjl}+\lambda_{cjl}}-\frac{1}{2}\Big(\frac{n_{cjl}}{m_{cjl}+\lambda_{cjl}}\Big)^2+\cdots\Big)\\
&\quad-\sum_{c=1}^{k}\sum_{j=1}^{d} n_c\log(n_c+m_c+1)\\
&\quad+\sum_{c=1}^{k}\sum_{j=1}^{d}\sum_{l=1}^{r_j} n_{cjl}\log(n_{cjl}+m_{cjl}+\lambda_{cjl}),
\end{aligned}
$$

where a series expansion of logarithm was used after the second equality sign, and thus we get

$$
\text{(A.2)}\quad \log p(\mathbf{x}^{(N)}|\mathbf{z}^{(M)},S,T)\approx
-\sum_{c=1}^{k}\sum_{j=1}^{d} n_c\log(n_c+m_c+1)+\sum_{c=1}^{k}\sum_{j=1}^{d}\sum_{l=1}^{r_j} n_{cjl}\log(n_{cjl}+m_{cjl}+\lambda_{cjl}),
$$

where it is assumed that $m_c$ is sufficiently large for all classes and that all limits of the relative frequencies of feature values are strictly positive under an infinitely exchangeable sampling process of the data, *i.e.*

$$
\frac{m_{cjl}}{m_c}\underset{m\to\infty}{\to}\beta_{cjl}>0,\quad c=1,\ldots,k,\quad j=1,\ldots,d,\quad l=1,\ldots,r_j.
$$

For the marginal predictive classifier MPC we can derive the comparable expression by the same expansion as (A.1) for

$$
\text{(A.3)}\qquad \sum_{c=1}^{k}\sum_{i:S_i=c}\log p(\mathbf{x}_i|\mathbf{z}^{(M)},T,S_i=c),
$$

which will give the result as

$$
\text{(A.4)}\quad
\begin{aligned}
&\sum_{c=1}^{k}\sum_{i:S_i=c}\log p(\mathbf{x}_i|\mathbf{z}^{(M)},T,S_i=c)\approx\\
&-\sum_{c=1}^{k}\sum_{j=1}^{d} n_c\log(m_c+2)+\sum_{c=1}^{k}\sum_{i:S_i=c}\sum_{j=1}^{d}\sum_{l=1}^{r_j} n_{i;cjl}\log(n_{i;cjl}+m_{cjl}+\lambda_{cjl}),
\end{aligned}
$$

where the sufficient statistic $n_{i;cjl}$ equals 1 if $\mathbf{x}_i$ carries the value $l$ for feature $j$ and $S_i=c$, and $n_{i;cjl}=0$ otherwise.

Consider now the relationship between data predictive probabilities for the two classifiers under the same classification structure, by investigation difference between (A.2) and (A.3), which can be expressed as

$$
\begin{aligned}
&\log p(\mathbf{x}^{(N)}|\mathbf{z}^{(M)}, S, T) - \sum_{c=1}^{k} \sum_{i:S_i=c} \log p(\mathbf{x}_i|\mathbf{z}^{(M)}, T, S_i = c) = \\
&- \sum_{c=1}^{k} \sum_{j=1}^{d} n_c \log(n_c + m_c + 1) \\
&+ \sum_{c=1}^{k} \sum_{j=1}^{d} n_c \log(m_c + 2) \\
&+ \sum_{c=1}^{k} \sum_{j=1}^{d} \sum_{l=1}^{r_j} n_{cjl} \log(n_{cjl} + m_{cjl} + \lambda_{cjl}) \\
&- \sum_{c=1}^{k} \sum_{i:S_i=c} \sum_{j=1}^{d} \sum_{l=1}^{r_j} n_{i;cjl} \log(n_{i;cjl} + m_{cjl} + \lambda_{cjl}) \\
= &- \sum_{c=1}^{k} \sum_{j=1}^{d} n_c \log\left(\frac{n_c + m_c + 1}{m_c + 2}\right) - \sum_{c=1}^{k} \sum_{j=1}^{d} \sum_{l=1}^{r_j} \log\left(\frac{n_{cjl} + m_{cjl} + \lambda_{cjl}}{1 + m_{cjl} + \lambda_{cjl}}\right)^{n_{cjl}} \underset{m\to\infty}{\to} 0.
\end{aligned}
$$

Thus, for any particular classification structure according to the joint rule, the log predictive probability of the data is asymptotically the same as the log predictive probability according to the corresponding marginal classifier, which implies that the two classifiers MPC and SPC will lead to equivalent classifications. □

## A.2 Theorem 2

*Proof.* We will establish the result by considering the difference of the logarithms of the two predictive probabilities and show that it converges to zero when $n \to \infty$. Firstly, the logarithm of the left hand side equals

$$\log p(\mathbf{x}^{(\delta)} | \mathbf{x}^{(N)}, s^{(\delta)}, \mathbf{z}^{(M)}, S, T) \tag{A.5}$$

$$= \sum_{c=1}^{k} \sum_{j=1}^{d} \log \frac{\Gamma(n_c + m_c + 1)}{\Gamma(\delta_c + n_c + m_c + 1)} + \sum_{c=1}^{k} \sum_{j=1}^{d} \sum_{l=1}^{\chi_j} \log \frac{\Gamma(\delta_{cjl} + n_{cjl} + m_{cjl} + \lambda_{cjl})}{\Gamma(n_{cjl} + m_{cjl} + \lambda_{cjl})},$$

where $\delta_c, \delta_{cjl}$ are the counterparts of the sufficient statistics $n_c, n_{cjl}$ from the additional test data, respectively. Using Stirling's approximation on the gamma functions in the above equation, we have

$$\log p(\mathbf{x}^{(\delta)} | \mathbf{x}^{(N)}, s^{(\delta)}, \mathbf{z}^{(M)}, S, T) \tag{A.6}$$

$$\approx - \sum_{c=1}^{k} \sum_{j=1}^{d} (n_c + m_c + \frac{1}{2}) \log(1 + \frac{\delta_c}{n_c + m_c + 1})$$
$$+ \sum_{c=1}^{k} \sum_{j=1}^{d} \sum_{l=1}^{\chi_j} (n_{cjl} + m_{cjl} + \lambda_{cjl} - \frac{1}{2}) \log(1 + \frac{\delta_{cjl}}{n_{cjl} + m_{cjl} + \lambda_{cjl}})$$
$$- \sum_{c=1}^{k} \sum_{j=1}^{d} \delta_c \log(\delta_c + n_c + m_c + 1)$$
$$+ \sum_{c=1}^{k} \sum_{j=1}^{d} \sum_{l=1}^{\chi_j} \delta_{cjl} \log(\delta_{cjl} + n_{cjl} + m_{cjl} + \lambda_{cjl}).$$

Further, using a series expansion of $\log(1 + x)$

$$= - \sum_{c=1}^{k} \sum_{j=1}^{d} (n_c + m_c + \frac{1}{2}) [\frac{\delta_c}{n_c + m_c + 1} - \frac{1}{2} (\frac{\delta_c}{n_c + m_c + 1})^2 + \dots]$$
$$+ \sum_{c=1}^{k} \sum_{j=1}^{d} \sum_{l=1}^{\chi_j} (n_{cjl} + m_{cjl} + \lambda_{cjl} - \frac{1}{2}) [\frac{\delta_{cjl}}{n_{cjl} + m_{cjl} + \lambda_{cjl}} - \frac{1}{2} (\frac{\delta_{cjl}}{n_{cjl} + m_{cjl} + \lambda_{cjl}})^2 + \dots]$$
$$- \sum_{c=1}^{k} \sum_{j=1}^{d} \delta_c \log(\delta_c + n_c + m_c + 1)$$
$$+ \sum_{c=1}^{k} \sum_{j=1}^{d} \sum_{l=1}^{\chi_j} \delta_{cjl} \log(\delta_{cjl} + n_{cjl} + m_{cjl} + \lambda_{cjl})$$

$$\approx -\sum_{c=1}^{k}\sum_{j=1}^{d}\delta_c \log(\delta_c + n_c + m_c + 1)$$

$$+\sum_{c=1}^{k}\sum_{j=1}^{d}\sum_{l=1}^{\chi_j}\delta_{cjl}\log(\delta_{cjl} + n_{cjl} + m_{cjl} + \lambda_{cjl}),$$

where a series expansion of $\log(1+x)$ was applied for the second equality. Thus, we obtain

$$\text{(A.7)} \qquad \log p(\mathbf{x}^{(\delta)}|\mathbf{x}^{(N)}, s^{(\delta)}, \mathbf{z}^{(M)}, S, T)$$

$$\approx -\sum_{c=1}^{k}\sum_{j=1}^{d}\delta_c \log(\delta_c + n_c + m_c + 1) + \sum_{c=1}^{k}\sum_{j=1}^{d}\sum_{l=1}^{\chi_j}\delta_{cjl}\log(\delta_{cjl} + n_{cjl} + m_{cjl} + \lambda_{cjl}),$$

where we use the assumption in (3.32). Similarly, log predictive probability of the labels of the $\delta$ additional items given the training and test data up to $n$ can be approximated as

$$\log p(s^{(\delta)}|\mathbf{x}^{(N)}, \mathbf{z}^{(M)}, S, T)$$

$$\approx -\delta \log(\delta + n + m + k) + \sum_{c=1}^{k}\delta_c \log(\delta_c + n_c + m_c + 1)$$

Thus, we obtain the approximate log predictive probability for both the labels and the additional test data

$$\text{(A.8)} \qquad \log p(\mathbf{x}^{(\delta)}, s^{(\delta)}|\mathbf{x}^{(N)}, \mathbf{z}^{(M)}, S, T)$$

$$\approx -\sum_{c=1}^{k}\sum_{j=1}^{d}\delta_c \log(\delta_c + n_c + m_c + 1) + \sum_{c=1}^{k}\sum_{j=1}^{d}\sum_{l=1}^{\chi_j}\delta_{cjl}\log(\delta_{cjl} + n_{cjl} + m_{cjl} + \lambda_{cjl})$$

$$-\delta \log(\delta + n + m + k) + \sum_{c=1}^{k}\delta_c \log(\delta_c + n_c + m_c + 1).$$

To develop a corresponding expansion for the right hand side of the theorem result, we can write

$$\text{(A.9)} \qquad \sum_{i=n+1}^{n+\delta}\log p(s_i, \mathbf{x}_i|\mathbf{x}^{(N)}, \mathbf{z}^{(M)}, S, T)$$

$$\approx -\sum_{c=1}^{k}\sum_{j=1}^{d}\delta_c \log(n_c + m_c + 1) + \sum_{c=1}^{k}\sum_{j=1}^{d}\sum_{l=1}^{\chi_j}\delta_{cjl}\log(n_{cjl} + m_{cjl} + \lambda_{cjl})$$

$$-\delta \log(n + m + k) + \sum_{c=1}^{k}\delta_c \log(n_c + m_c + 1).$$

We now obtain the stated results by considering the difference between (A.8) and (A.9), which satisfies

$$\sum_{c=1}^{k}\sum_{j=1}^{d}\log(\frac{n_c+m_c+1}{\delta_c+n_c+m_c+1})^{\delta_c}+\sum_{c=1}^{k}\sum_{j=1}^{d}\sum_{l=1}^{\chi_j}\log(\frac{n_{cjl}+m_{cjl}+\lambda_{cjl}}{\delta_{cjl}+n_{cjl}+m_{cjl}+\lambda_{cjl}})^{\delta_{cjl}}$$

$$+\log(\frac{n+m+k}{\delta+n+m+k})^{\delta}+\sum_{c=1}^{k}\log(\frac{n_c+m_c+1}{\delta_c+n_c+m_c+1})^{\delta_c}\underset{n\to\infty}{\to}0,$$

where the last step necessitates use of the condition stated in (3.32).

□

## A.3 Corollary 1

*Proof.* We have above established the asymptotic equality between the posterior probabilities of the label $s_t$ given $\mathbf{x}_t$ and either only the earlier history of the process up to $n$, or both all the new test data $\mathbf{x}^{(\delta)}$ and the earlier history. In other words, we have that

$$p(s_t|\mathbf{x}^{(N)}, \mathbf{x}^{(\delta)}, \mathbf{z}^{(M)}, S, T) \underset{n\to\infty}{\to} p(s_t|\mathbf{x}^{(N)}, \mathbf{x}_t, \mathbf{z}^{(M)}, S, T). \tag{A.10}$$

Then, given a sufficient long history of test items, marginalization over subsequential items cannot induce any changes in the marginal posterior. This can be concluded by first considering

$$p(s_t|\mathbf{x}^{(N)}, \mathbf{x}^{(\delta)}, \mathbf{z}^{(M)}, S, T) = \sum_{s^{(\delta)}\setminus\{t\}} p(s^{(\delta)}|\mathbf{x}^{(N)}, \mathbf{x}^{(\delta)}, \mathbf{z}^{(M)}, S, T), \tag{A.11}$$

and then, by considering

$$p(s_t|\mathbf{x}^{(N)}, \mathbf{x}_t, \mathbf{z}^{(M)}, S, T) = \sum_{S, s^{(\delta)}\setminus\{t\}} \prod_{i=n+1}^{n+\delta} p(s_i|\mathbf{x}^{(N)}, \mathbf{x}_i, \mathbf{z}^{(M)}, S, T). \tag{A.12}$$

Thus, the asymptotic equality of (A.11) and (A.12) stated above implies the equality of the two classifiers.

□

## A.4 Lemma 1

*Proof.* The logarithm $\log p(\mathbf{x}^{(N)}|\mathbf{z}^{(M)}, S, T)$ of the predictive probability equals

$$\sum_{c=1}^{k}\sum_{j=1}^{d}\{\log[(m_c+n_c)!]$$
$$-\log(m_c!) - \log[(\psi+m_c)(\psi+m_c+1)\cdots(\psi+m_c+n_c-1)]$$
$$+\sum_{t=1}^{m_c+n_c}\tilde{\rho}_{cjt}\log(\frac{\psi}{t}) - \sum_{t=1}^{m_c}\rho_{cjt}\log(\frac{\psi}{t}) - \sum_{t=1}^{m_c+n_c}\log(\tilde{\rho}_{cjt}!) + \sum_{t=1}^{m_c}\log(\rho_{cjt}!)\}.$$

Correspondingly, the logarithm for the marginal classifier is

$$\log p(\mathbf{x}^{(N)}|\mathbf{z}^{(M)}, \mathring{S}, T) = \sum_{c=1}^{k}\sum_{i\in s_c}\sum_{j=1}^{d}\{\log[(m_c+1)!] - \log(m_c!) - \log(\psi+m_c)$$
$$+\sum_{t=1}^{m_c+1}\tilde{\rho}_{cjt}{}^{(i)}\log(\frac{\psi}{t}) - \sum_{t=1}^{m_c}\rho_{cjt}\log(\frac{\psi}{t}) - \sum_{t=1}^{m_c+1}\log(\tilde{\rho}_{cjt}{}^{(i)}!) + \sum_{t=1}^{m_c}\log(\rho_{cjt}!)\Bigg\}.$$

To arrive at the stated result, we will first show that the differences of the foremost terms in the two expressions of the log probabilities tend to 0 as $m_c$ increases. To start, recall the Ramanujan equation for log factorial, which is given by

$$\log(a!) = a\log a - a + \frac{1}{6}\log\{a + 4a^2 + 8a^3 + \epsilon(a)/30\} + \frac{1}{2}\log\pi,$$

where the error term $\epsilon(a)$ is bounded as $3/10 < \epsilon(a) < 1$ and $\epsilon(a) \to 1$ when $a \to \infty$. It then follows that

$$\log[(m_c+n_c)!] - \log(m_c)! - \sum_{i\in s_c}\log[(m_c+1)!] + \sum_{i\in s_c}\log(m_c!)$$
$$\approx (m_c+n_c)\log(m_c+n_c) - (m_c+n_c)$$
$$+\frac{1}{6}\log[(m_c+n_c) + 4(m_c+n_c)^2 + 8(m_c+n_c)^3] - m_c\log(m_c) + m_c$$
$$-\frac{1}{6}\log(m_c + 4{m_c}^2 + 8{m_c}^3) + n_c m_c\log(m_c) - n_c m_c$$
$$+\frac{n_c}{6}\log(m_c + 4{m_c}^2 + 8{m_c}^3) - n_c(m_c+1)\log(m_c+1) + n_c(m_c+1)$$
$$-\frac{n_c}{6}\log[(m_c+1) + 4(m_c+1)^2 + 8(m_c+1)^3]$$

$$= m_c \log(1 + \frac{n_c}{m_c}) - m_c n_c \log(1 + \frac{1}{m_c}) + n_c \log(1 + \frac{n_c - 1}{m_c + 1})$$
$$+ \frac{1}{6}\log[\frac{(m_c + n_c) + 4(m_c + n_c)^2 + 8(m_c + n_c)^3}{m_c + 4m_c + 8m_c}]$$
$$- \frac{n_c}{6}\log[\frac{(m_c + 1) + 4(m_c + 1)^2 + 8(m_c + 1)^3}{m_c + 4{m_c}^2 + 8{m_c}^3}].$$

Using the standard series expansion $\log(1 + y) = y - \frac{1}{2}y^2 + \frac{1}{3}y^3 - \cdots$, the result above can be further simplified to the expression

$$\log[(m_c + n_c)!] - \log(m_c!) - \sum_{i \in s_c} \log[(m_c + 1)!] + \sum_{i \in s_c} \log(m_c!)$$
$$= m_c[\frac{n_c}{m_c} - \frac{1}{2}(\frac{n_c}{m_c})^2 + \cdots]$$
$$- m_c n_c[\frac{1}{m_c} - \frac{1}{2}(\frac{1}{m_c})^2 + \cdots]$$
$$+ n_c[\frac{n_c - 1}{m_c + 1} - \frac{1}{2}(\frac{n_c - 1}{m_c + 1})^2 + \cdots]$$
$$+ \frac{1}{6}\log\{\frac{(m_c + n_c) + 4(m_c + n_c)^2 + 8(m_c + n_c)^3}{m_c + 4{m_c}^2 + 8{m_c}^3}\}$$
$$- \frac{n_c}{6}\log\{\frac{(m_c + 1) + 4(m_c + 1)^2 + 8(m_c + 1)^3}{m_c + 4{m_c}^2 + 8{m_c}^3}\},$$

which tends to 0 when $m_c$ increases. For the difference of the intermediate terms in the log predictive probabilities a similar convergence result holds such that

$$- \log[(\psi + m_c)(\psi + m_c + 1) \cdots (\psi_{cj} + m_c + n_c - 1)] + \sum_{i \in s_c} \log(\psi + m_c)$$
$$= \log \frac{(\psi + m_c)^{n_c}}{(\psi + m_c)(\psi + m_c + 1) \cdots (\psi + m_c + n_c - 1)} \underset{m_c \to \infty}{\to} 0.$$

Note that the order of the error terms is not explicitly included in the above derivations to avoid too extensive and tedious expressions. To establish the remaining parts of the lemma, note first that ${\tilde{\rho}_{cjt_j^{(i)}}}^{(i)} = \rho_{cjt_j^{(i)}} - 1$ and ${\tilde{\rho}_{cj(t_j^{(i)}+1)}}^{(i)} =$ $\rho_{cj(t_j^{(i)}+1)} + 1$. Then, we obtain that $\sum_{t=1}^{m_c+1} \tilde{\rho}_{cjt}{}^{(i)} \log(\frac{\psi}{t})$ equals

$$\sum_{t=1}^{m_c} \Big[\rho_{cjt} \log(\frac{\psi}{t}) - \rho_{cjt_j^{(i)}} \log(\frac{\psi}{t_j^{(i)}}) - \rho_{cj(t_j^{(i)}+1)} \log(\frac{\psi}{t_j^{(i)}+1})$$

$$+ \tilde{\rho}_{cjt_j^{(i)}}{}^{(i)} \log(\frac{\psi}{t_j^{(i)}}) + \tilde{\rho}_{cj(t_j^{(i)}+1)}{}^{(i)} \log(\frac{\psi}{t_j^{(i)}+1})\Big]$$

$$= \sum_{t=1}^{m_c} \rho_{cjt} \log(\frac{\psi}{t}) - \log(\frac{\psi}{t_j^{(i)}}) + \log(\frac{\psi}{t_j^{(i)}+1}) = \sum_{t=1}^{m_c} \rho_{cjt} \log(\frac{\psi}{t}) - \log(\frac{t_j^{(i)}+1}{t_j^{(i)}}).$$

By using the same relationships between the original and updated sufficient statistics, we have that $\sum_{t=1}^{m_c+1} \log(\tilde{\rho}_{cjt}{}^{(i)}!)$ equals

$$\sum_{t=1}^{m_c} \log(\rho_{cjt}!) - \log(\rho_{cjt_j^{(i)}}!) - \log(\rho_{cj(t_j^{(i)}+1)}!) + \log(\tilde{\rho}_{cjt_j^{(i)}}{}^{(i)}!) + \log(\tilde{\rho}_{cj(t_j^{(i)}+1)}{}^{(i)}!)$$

$$= \sum_{t=1}^{m_c} \log(\rho_{cjt}!) - \log(\rho_{cjt_j^{(i)}}) + \log(\rho_{cj(t_j^{(i)}+1)} + 1).$$

Using these two results, the stated lemma follows after some tedious re-arrangements of the terms.

□

## A.5 Theorem 7

*Proof.* To arrive at the stated result, it is necessary to show that a positive probability remains for the event that the two classifiers disagree, despite of the amount of training data present. To demonstrate this, we will consider log ratios of the predictive probabilities under two particular classification structures and show that their difference need not converge to zero. Let $S = \mathring{S}$ be a labelling of $n$ test items and $S'$ the labelling derived from $S$ by re-assigning a single item $i$ from class $c_1$ to class $c_2$. Let $n'_{\mathbf{c}_1} = n_{\mathbf{c}_1} - 1$, $n'_{\mathbf{c}_2} = n_{\mathbf{c}_2} + 1$, and let $\tilde{\rho}'_{\mathbf{c}_1 jt}, \tilde{\rho}'_{\mathbf{c}_2 jt}$ denote the updated sufficient statistics under $S'$. Since the two labellings $S, S'$ are identical apart from the classes $c_1, c_2$, log ratio of the predictive probabilities $\log p(\mathbf{x}^{(N)}|\mathbf{z}^{(M)}, S, T) - \log p(\mathbf{x}^{(N)}|\mathbf{z}^{(M)}, S', T)$ under the simultaneous classifier can be written as

$$
\begin{aligned}
&\log p(\mathbf{x}^{(s_{c_1})}, \mathbf{z}^{(M)}|S, T) + \log p(\mathbf{x}^{(s_{c_2})}, \mathbf{z}^{(M)}|S, T) \\
&- \log p(\mathbf{x}^{(s_{c_1})}, \mathbf{z}^{(M)}|S', T) - \log p(\mathbf{x}^{(s_{c_2})}, \mathbf{z}^{(M)}|S', T) \\
&= \sum_{j=1}^{d} \Bigg\{ \Bigg[ \log \frac{(m_{\mathbf{c}_1} + n_{\mathbf{c}_1})!}{\psi(\psi+1)\cdots(\psi + m_{\mathbf{c}_1} + n_{\mathbf{c}_1} - 1)} + \sum_{t=1}^{m_{\mathbf{c}_1}+n_{\mathbf{c}_1}} \tilde{\rho}_{\mathbf{c}_1 jt} \log(\frac{\psi}{t}) + \sum_{t=1}^{m_{\mathbf{c}_1}+n_{\mathbf{c}_1}} \log \frac{1}{\tilde{\rho}_{\mathbf{c}_1 jt}!} \Bigg] \\
&+ \Bigg[ \log \frac{(m_{\mathbf{c}_2} + n_{\mathbf{c}_2})!}{\psi(\psi+1)\cdots(\psi + m_{\mathbf{c}_2} + n_{\mathbf{c}_2} - 1)} + \sum_{t=1}^{m_{\mathbf{c}_2}+n_{\mathbf{c}_2}} \tilde{\rho}_{\mathbf{c}_2 jt} \log(\frac{\psi}{t}) + \sum_{t=1}^{m_{\mathbf{c}_2}+n_{\mathbf{c}_2}} \log \frac{1}{\tilde{\rho}_{\mathbf{c}_2 jt}!} \Bigg] \\
&- \Bigg[ \log \frac{(m_{\mathbf{c}_1} + n'_{\mathbf{c}_1})!}{\psi(\psi+1)\cdots(\psi + m_{\mathbf{c}_1} + n'_{\mathbf{c}_1} - 1)} + \sum_{t=1}^{m_{\mathbf{c}_1}+n'_{\mathbf{c}_1}} \tilde{\rho}'_{\mathbf{c}_1 jt} \log(\frac{\psi}{t}) + \sum_{t=1}^{m_{\mathbf{c}_1}+n'_{\mathbf{c}_1}} \log \frac{1}{\tilde{\rho}'_{\mathbf{c}_1 jt}!} \Bigg] \\
&- \Bigg[ \log \frac{(m_{\mathbf{c}_2} + n'_{\mathbf{c}2})!}{\psi(\psi+1)\cdots(\psi + m_{\mathbf{c}_2} + n'_{\mathbf{c}2} - 1)} + \sum_{t=1}^{m_{\mathbf{c}_2}+n'_{\mathbf{c}2}} \tilde{\rho}'_{\mathbf{c}_2 jt} \log(\frac{\psi}{t}) + \sum_{t=1}^{m_{\mathbf{c}_2}+n'_{\mathbf{c}2}} \log \frac{1}{\tilde{\rho}'_{\mathbf{c}_2 jt}!} \Bigg] \Bigg\}.
\end{aligned}
$$

Given the simple relationships between the sufficient statistics, the above expression can be further simplified to

$$\sum_{j=1}^{d}\{\log(m_{\mathbf{c}_1}+n_{\mathbf{c}_1})-\log(m_{\mathbf{c}_2}+n_{\mathbf{c}_2}+1)-\log(\psi+m_{\mathbf{c}_1}+n_{\mathbf{c}_1}-1)+\log(\psi+m_{\mathbf{c}_2}+n_{\mathbf{c}_2})$$

$$+\left[\sum_{t=1}^{m_{\mathbf{c}_1}+n_{\mathbf{c}_1}}\tilde{\rho}_{\mathbf{c}_1jt}\log(\frac{\psi}{t})-\sum_{t=1}^{m_{\mathbf{c}_1}+n'_{\mathbf{c}_1}}\tilde{\rho}'_{\mathbf{c}_1jt}\log(\frac{\psi}{t})\right]+\left[\sum_{t=1}^{m_{\mathbf{c}_2}+n_{\mathbf{c}_2}}\tilde{\rho}_{\mathbf{c}_2jt}\log(\frac{\psi}{t})-\sum_{t=1}^{m_{\mathbf{c}_2}+n'_{\mathbf{c}_2}}\tilde{\rho}'_{\mathbf{c}_2jt}\log(\frac{\psi}{t})\right]$$

$$-\left[\sum_{t=1}^{m_{\mathbf{c}_1}+n_{\mathbf{c}_1}}\log(\tilde{\rho}_{\mathbf{c}_1jt}!)-\sum_{t=1}^{m_{\mathbf{c}_1}+n'_{\mathbf{c}_1}}\log(\tilde{\rho}'_{\mathbf{c}_1jt}!)\right]-\left[\sum_{t=1}^{m_{\mathbf{c}_2}+n_{\mathbf{c}_2}}\log(\tilde{\rho}_{\mathbf{c}_2jt}!)-\sum_{t=1}^{m_{\mathbf{c}_2}+n'_{\mathbf{c}_2}}\log(\tilde{\rho}'_{\mathbf{c}_2jt}!)\right]\Bigg\}.$$

Assume now that item $i$ carries a unique value for feature $j$ and no other unique values are present in classes $c_1, c_2$ for this feature. This event has a strictly positive probability [53]. Then, the vectors of sufficient statistics $\tilde{\rho}_{\mathbf{c}_1jt}, \tilde{\rho}'_{\mathbf{c}_1jt}$ are identical apart from the first element, such that $\tilde{\rho}_{\mathbf{c}_1j1}=1, \tilde{\rho}'_{\mathbf{c}_1j1}=0$. Consequently, the difference in the first square bracket equals $\log(\psi)$. Similarly, we obtain for the second square bracket: $\tilde{\rho}_{\mathbf{c}_2j1}=0, \tilde{\rho}'_{\mathbf{c}_2j1}=1$, and thus, the difference equals $-\log(\psi)$. Using a similar argument, differences in both remaining two square brackets equal zero. Finally, with an increasing $m_c$ the log ratio of predictive probabilities then converges to zero under the simultaneous classifier. This can be interpreted as an intuitive result, since the surprising event observed for item $i$ is equally surprising in both populations represented by $s_{c_1}, s_{c_2}$ and the corresponding training data sets. Consider now the log ratio of predictive probabilities under the marginal classifier. Since the predictive probabilities of all items apart from $i$ are identical, the log ratio simplifies to $\log p(\mathbf{x}_i, \mathbf{z}^{(M)}|i\in s_{\mathbf{c}_1}, T)-\log p(\mathbf{x}_i, \mathbf{z}^{(M)}|i\in s_{\mathbf{c}_2}, T)$ which equals

$$\sum_{j=1}^{d}\left\{\left[\log\frac{(m_{\mathbf{c}_1}+1)!}{\psi(\psi+1)\cdots(\psi+m_{\mathbf{c}_1}+1-1)}+\sum_{t=1}^{m_{\mathbf{c}_1}+1}\tilde{\rho}_{\mathbf{c}_1jt}{}^{(i)}\log(\frac{\psi}{t})-\sum_{t=1}^{m_{\mathbf{c}_1}+1}\log(\tilde{\rho}_{\mathbf{c}_1jt}{}^{(i)}!)\right]\right.$$

$$\left.-\left[\log\frac{(m_{\mathbf{c}_2}+1)!}{\psi(\psi+1)\cdots(\psi+m_{\mathbf{c}_2}+1-1)}+\sum_{t=1}^{m_{\mathbf{c}_2}+1}\tilde{\rho}_{\mathbf{c}_2jt}{}^{(i)}\log(\frac{\psi}{t})-\sum_{t=1}^{m_{\mathbf{c}_2}+1}\log(\tilde{\rho}_{\mathbf{c}_2jt}{}^{(i)}!)\right]\right\}.$$

As for the simultaneous classifier, difference of the foremost terms not involving the sufficient statistics converges to zero when $m_c$ increases. However, here the difference of the latter terms can be arbitrarily large, as it depends basically on the difference between the sufficient statistics derived from the training data for the two classes. Thus, there is a positive probability for the event that the two classifiers disagree, even if the difference of log ratios would converge to zero for the remaining $d-1$ features. □

# Appendix B

# Some Mathematical Functions

## B.1 Gamma and beta functions

Gamma and beta functions are special functions which are needed for the normalizing constants of some of the standard distributions.

**Gamma** function can be defined by the integral

$$\Gamma(z) = \int_0^\infty x^{z-1} e^{-x} \, dx, \qquad z > 0.$$

It satisfies the functional equation

$$\Gamma(z+1) = z\,\Gamma(z), \qquad \text{for all } z > 0,$$

and besides $\Gamma(1) = 1$, from which it follows that

$$\Gamma(n) = (n-1)!, \qquad \text{when } n = 1, 2, 3, \dots .$$

Therefore the gamma function is a generalization of the factorial. The value of, $\Gamma(z)$ for half-integer arguments can be calculated using its functional equation and the value $\Gamma(\frac{1}{2}) = \sqrt{\pi}$.

**Beta** function can be defined by the integral

$$B(a, b) = \int_0^1 u^{a-1}(1-u)^{b-1} \, du, \qquad a, b > 0.$$

It has the following connection with the gamma function,

$$B(a, b) = \frac{\Gamma(a)\Gamma(b)}{\Gamma(a+b)}.$$

## B.2 Stirling's approximation to gamma function

In mathematics Stirling's approximation (or Stirling's formula) is an approximation for large factorials that is due to James Stirling. The formula as typically used in application is

$$\ln x! = x \ln x - x + O(\ln(x)),$$

where the next term of the $O(\ln(x))$ is

$$O(\ln(x)) \approx \frac{1}{2} \ln(2\pi x).$$

The alternative form of this formula is therefore

$$\lim_{x \to \infty} \frac{x!}{\sqrt{2\pi x}\left(\frac{x}{e}\right)^x} = 1.$$

That is often written as

$$x! \approx \left(\frac{x}{e}\right)^x \sqrt{2\pi x}.$$

## B.3 Series expansion of a function

A series expansion is a representation of a particular function as a sum of powers in one of its variables, or by a sum of powers of another (usually elementary) function $f(x)$ and a Maclaurin series is a Taylor series expansion of a function about 0,

$$f(x) = f(0) + f^{'}(0)x + \frac{f^{''}(0)}{2!}x^2 + \frac{f^{(3)}(0)}{3!}x^3 + \ldots + \frac{f^{(n)}(0)}{n!}x^n + \ldots.$$

Maclaurin series are named after the Scottish mathematician Colin Maclaurin. In particular if we consider the $f(x)$ as $\ln(1+x)$ its Maclaurin series expansion would be

$$\ln(1+x) = x - \frac{1}{2}x^2 + \frac{1}{3}x^3 - \frac{1}{4}x^4 + \ldots,$$

that is the same as its explicit compact form of

$$\ln(1+x) = \sum_{n=1}^{\infty} \frac{(-1)^{n+1}}{n} x^n.$$

# Appendix C

# Distributions

## C.1 Univariate discrete distributions

**Binomial** distribution $\mathrm{Bin}(n, p)$, $n$ positive integer, $0 \le p \le 1$, has pmf

$$\mathrm{Bin}(x \mid n, p) = \binom{n}{x} p^x (1-p)^{n-x}, \quad x = 0, 1, \ldots, n.$$

If $X \sim \mathrm{Bin}(n, p)$, then

$$EX = np, \qquad \operatorname{var} X = np(1-p).$$

When $n = 1$, the binomial is also called the **Bernoulli** distribution.

**Poisson** distribution $\mathrm{Poi}(\theta)$ with parameter $\theta > 0$ has pmf

$$\mathrm{Poi}(x \mid \theta) = \mathrm{e}^{-\theta} \frac{\theta^x}{x!}, \quad x = 0, 1, 2, \ldots$$

If $X \sim \mathrm{Poi}(\theta)$, then

$$EX = \theta, \qquad \operatorname{var} X = \theta.$$

## C.2 Univariate continuous distributions

**Beta** distribution $\mathrm{Be}(a, b)$ with parameters $a > 0, b > 0$ has pdf

$$\mathrm{Be}(x \mid a, b) = \frac{1}{B(a, b)} x^{a-1} (1-x)^{b-1}, \quad 0 < x < 1.$$

$B(a, b)$ is the beta function with arguments $a$ and $b$. If $X \sim \mathrm{Be}(a, b)$, then

$$EX = \frac{a}{a+b}, \qquad \operatorname{var} X = \frac{ab}{(a+b)^2\,(a+b+1)}.$$

The uniform distribution $\mathrm{Uni}(0,1)$ is the same as $\mathrm{Be}(1,1)$.

**Chi squared** distribution $\chi^2_\nu$ with $\nu > 0$ degrees of freedom is the same as the gamma distribution

$$\mathrm{Gam}(\frac{\nu}{2}, \frac{1}{2}).$$

If $X \sim \chi^2_\nu$, then $EX = \nu$ and $\operatorname{var} X = 2\nu$.

**Exponential** distribution $\mathrm{Exp}(\lambda)$ with rate $\lambda > 0$ has pdf

$$\mathrm{Exp}(x \mid \lambda) = \lambda\, \mathrm{e}^{-\lambda x}, \quad x > 0.$$

If $X \sim \mathrm{Exp}(\lambda)$, then

$$EX = \frac{1}{\lambda}, \qquad \operatorname{var} X = \frac{1}{\lambda^2}.$$

**Gamma** distribution $\mathrm{Gam}(a, b)$ with parameters $a > 0, b > 0$ has pdf

$$\mathrm{Gam}(x \mid a, b) = \frac{b^a}{\Gamma(a)}\, x^{a-1} \mathrm{e}^{-bx}, \quad x > 0.$$

$\Gamma(a)$ is the gamma function. If $X \sim \mathrm{Gam}(a, b)$, then

$$EX = \frac{a}{b}, \qquad \operatorname{var} X = \frac{a}{b^2}.$$

Exponential distribution $\mathrm{Exp}(\lambda)$ is the same as $\mathrm{Gam}(1, \lambda)$.

**Uniform** distribution $\mathrm{Uni}(a, b)$ on the interval $(a, b)$, where $a < b$, has pdf

$$\mathrm{Uni}(x \mid a, b) = \frac{1}{b-a}, \qquad a < x < b.$$

If $X \sim \mathrm{Uni}(a, b)$, then

$$EX = \frac{1}{2}(a+b), \qquad \operatorname{var} X = \frac{1}{12\,(b-a)^2}.$$

## C.3 Multivariate discrete distributions

**Multinomial** distribution $\mathrm{Mult}(n, (p_1, p_2, \ldots, p_k))$ with sample size $n$ and probability vector parameter $(p_1, \ldots, p_k)$ has pmf

$$\mathrm{Mult}(x_1, \ldots, x_k \mid n, (p_1, \ldots, p_k)) = \frac{n!}{\prod_{i=1}^k x_i!} \prod_{j=1}^k p_j^{x_j},$$

when $x_1, \dots, x_k \geq 0$ are integers summing to $n$ (and the pmf is zero otherwise). If $X \sim \text{Mult}(n, p)$, then $EX_i = np_i$, $\operatorname{var} X_i = np_1(1 - p_i)$, and $\operatorname{cov}(X_i, X_j) = -np_i p_j$, when $i \neq j$.

If $(X, Y) \sim \text{Mult}(n, (p, 1 - p))$, then $X \sim \text{Bin}(n, p)$.

## C.4 Multivariate continuous distributions

**Dirichlet** distribution $\text{Dir}(a_1, \dots, a_{d+1})$ with parameters $a_1, \dots, a_{d+1} > 0$ is the $d$-dimensional distribution with the pdf

$$\text{Dir}(x \mid a) = \text{Dir}(x \mid a_1, \dots, a_{d+1}) = \frac{\Gamma(a_1 + \dots + a_{d+1})}{\Gamma(a_1) \cdots \Gamma(a_{d+1})} x_1^{a_1 - 1} x_2^{a_2 - 1} \dots x_d^{a_d - 1} (1 - x_1 - x_2 - \dots - x_d)^{a_{d+1} - 1},$$

when

$$x_1, \dots, x_d > 0, \quad \text{and} \quad x_1 + \dots + x_d < 1,$$

and zero otherwise. If $X \sim \text{Dir}(a_1, \dots, a_{d+1})$, and $s = \sum a_i$, then

$$EX_i = \frac{a_i}{s}, \quad \operatorname{var} X_i = \frac{a_i(s - a_i)}{s^2(s + 1)}, \quad \operatorname{cov}(X_i, X_j) = \frac{-a_i a_j}{s^2(s + 1)}, \quad \text{when } i \neq j.$$

The univariate Dirichlet $\text{Dir}(a_1, a_2)$ is the same as the beta distribution $\text{Be}(a_1, a_2)$.